\documentclass[final,5p,times,twocolumn]{elsarticle}

\usepackage{amssymb}
\usepackage{amsmath}
\usepackage{cite}
\usepackage[linesnumbered,ruled,vlined]{algorithm2e}
\usepackage{booktabs}
\usepackage{graphicx}
\usepackage{textcomp}
\usepackage{xcolor}
\usepackage{color}
\usepackage[binary-units]{siunitx}
\usepackage{multicol}
\usepackage{multirow}
\usepackage{bm}
\usepackage{subcaption}
\usepackage{url}
\usepackage{autobreak}
\usepackage[para]{footmisc}
\usepackage{makecell}
\usepackage{xspace}
\usepackage{comment}
\usepackage{eso-pic}

\AddToShipoutPictureFG*{%
\AtPageUpperLeft{%
\raisebox{-1.5cm}[0pt][0pt]{%
\makebox[\paperwidth]{%
\hfill
\textcolor{red}{This is the first submitted version of the manuscript.}
\hfill
}}}}

\newcommand{\Malaga}{M\'{a}laga\xspace}
\expandafter\def\expandafter\normalsize\expandafter{%
    \normalsize%
    \setlength\abovedisplayskip{2pt}%
    \setlength\belowdisplayskip{2pt}%
    \setlength\abovedisplayshortskip{2pt}%
    \setlength\belowdisplayshortskip{2pt}%
}

\journal{Future Generation Computer Systems}

\begin{document}

\begin{frontmatter}



\title{Green Optimization: Energy-aware Design of Metaheuristics by Using Machine Learning Surrogates to Cope with Real Problems}

\author[label1]{Tomohiro Harada\corref{cor1}}
\ead{tharada@mail.saitama-u.ac.jp}
\author[label2]{Enrique Alba}
\ead{eat@lcc.uma.es}
\author[label2]{Gabriel Luque}
\ead{gluque@uma.es}

\affiliation[label1]{organization={Graduate School of Science and Engineering, Saitama University},
            state={Saitama},
            country={Japan}}

\affiliation[label2]{organization={ITIS Software, University of Malaga},
            state={Malaga},
            country={Spain}}



\begin{abstract}
Addressing real-world optimization challenges requires not only advanced metaheuristics but also continuous refinement of their internal mechanisms. This paper explores the integration of machine learning in the form of neural surrogate models into metaheuristics through a recent lens: energy consumption. While surrogates are widely used to reduce the computational cost of expensive objective functions, their combined impact on energy efficiency, algorithmic performance, and solution accuracy remains largely unquantified. We provide a critical investigation into this intersection, aiming to advance the design of energy-aware, surrogate-assisted search algorithms. Our experiments reveal substantial benefits: employing a state-of-the-art pre-trained surrogate can reduce energy consumption by up to 98\%, execution time by approximately 98\%, and memory usage by around 99\%. Moreover, increasing the training dataset size further enhances these gains by lowering the per-use computational cost, while static pre-training versus continuous (iterative) retraining have relatively different advantages depending on whether we aim at time/energy or accuracy and general cost across problems, respectively. Surrogates also have a negative impact on costs and accuracy at times, and then they cannot be blindly adopted. These findings support a more holistic approach to surrogate-assisted optimization, integrating energy with time and predictive accuracy into performance assessments.
\end{abstract}



\begin{keyword}


green computing \sep profiling energy of algorithms \sep surrogates for fitness evaluation \sep real problems
\end{keyword}

\end{frontmatter}



\section{Introduction}
Solving real-world problems is inherently complex, involving stages such as problem modeling, user interaction, and algorithmic implementation. Contemporary research in problem-solving methodologies emphasizes achieving both efficiency and precision~\citep{OSABA2021100888}. Typically, efficiency is assessed in terms of computational runtime~\citep{alba2002improving, ABDELHAFEZ2020100692}, often overlooking other critical factors such as memory usage~\citep{khalfi2023metaheuristics}. Regarding accuracy, many works evaluate the deviation from known optima or reference solution costs as the primary metric for solver quality~\citep{He2016Convergence}. While diverse techniques exist to fulfill these basic objectives~\citep{Halim2021}, a clear gap remains in conducting more holistic evaluations of algorithms, where significant operational dimensions are frequently neglected.

Among these overlooked dimensions, the energy profile of computational methods stands out as a crucial factor. Although preliminary investigations have begun to consider energy consumption in algorithms~\citep{Abdelhafez2019,Jamil2022}, a broader understanding is still lacking. The implications of energy-aware computing span multiple axes: the practical deployability of algorithms, real cost efficiency in high-performance computing, and compliance with the principles of Green AI~\citep{Paul2023Green}. In today’s context, where ecological concerns intersect with technological advancement, low-power, energy-efficient algorithmic designs are no longer optional but essential.

One prominent strategy for enhancing the efficiency of metaheuristics in real problems involves the use of surrogate models. In many real-world applications, the objective function—frequently a detailed simulation of the system under study—imposes high computational and memory demands. Surrogates such as neural networks (NNs)~\citep{iima2023genetic} offer a compelling alternative by acting as computationally inexpensive proxies that approximate the behavior of the true fitness function. These models allow the solver to explore the solution space more rapidly while preserving guidance toward promising regions of the search landscape.

Moreover, the current literature on surrogate-assisted optimization predominantly concentrates on reducing computational time~\citep{Li2022PHEV}, often disregarding the overhead associated with training surrogate models. The prevailing assumption is that any significant decrease in time-to-solution justifies the surrogate's utility~\citep{TONG2021414}. Yet, in pursuit of a more principled algorithm design, it becomes necessary to adopt a more nuanced approach---one that rigorously interrogates not only the performance gains but also the energy cost, reliability, and overall efficiency of surrogate integration. This study seeks to advance that direction by posing the following research questions:

\begin{description}
    \setlength\parskip{-1em}%
    \setlength\itemsep{0em}%
\item[RQ1] — Surrogate Energy Cost: How much energy does the surrogate consume, and how does it compare to the original fitness function and overall solver?
    \setlength\parskip{0em}%
\item[RQ2] — Redefining Efficiency: With energy profiles for fitness, surrogate, and solver, how do we now define algorithmic efficiency and accuracy?
\item[RQ3] — Training Overhead: Is a fully trained surrogate necessary at all, or is fast, low-cost training sufficient for effective optimization?
\item[RQ4] — Static vs. Dynamic Surrogates: Is it better to pre-train and fix the surrogate, or to refine it iteratively during the metaheuristic search?
\end{description}
\vspace{-1em}

These questions point to a more global methodology for surrogate-assisted algorithms, where not only conventional metrics like time, number of evaluations, or success of the solver are included, but also the energy profile and the numerical accuracy of the surrogate model are both taken into account. By answering the previous questions, we will foster new thinking on when and how surrogates are useful, and we will provide some initial insights for future comparisons.

In this article, we explore surrogate-assisted optimization using multiple search templates, specifically particle swarm optimization (PSO) and genetic algorithms (GA), applied to a real mobility problem in smart cities (optimal tuning of traffic light durations for efficient road traffic). Additionally, we consider different surrogate integration strategies, including pre-trained and iteratively updated NN-based models, to assess their impact on efficiency and energy consumption. While these choices were made to ensure a focused and structured analysis, our findings are potentially applicable to other optimization techniques, problem domains, and surrogate-assisted search frameworks.


This work begins with a review of relevant works in Section~\ref{sec:related_work}, while Section~\ref{sec:problem} describes the traffic light scheduling problem. We then discuss surrogate-assisted optimization and different ways of designing PSO and GA (Section~\ref{sec:surrogate_methods}). Experiments and result evaluation are detailed in Section~\ref{sec:experimental_setup}, followed by an analysis of computational resource consumption at different stages of the process: evaluating candidate solutions using SUMO (Section~\ref{sec:solution_evaluation}), training the surrogate model (Section~\ref{sec:nn_training}), and using it within the resulting algorithms (Section~\ref{sec:algorithm_execution}). Finally, Section~\ref{sec:conclusions} summarizes our main findings and discusses future research directions.

\section{Related Work}\label{sec:related_work}
Research on energy efficiency and green computing in AI, including machine learning, has gained attention~\citep{Verdecchia2023}. This section reviews prior studies from two perspectives: the energy consumption of search algorithms and the efficiency and accuracy of surrogate-assisted methods.

\subsection{Energy Consumption of Search Algorithms}

Recent studies have assessed the energy characteristics of evolutionary algorithms (EAs) and metaheuristics in various contexts. Abdelhafez et al. examined the energy usage of sequential and parallel genetic algorithms at the component level~\citep{Abdelhafez2019}. Jamil et al. compared the energy efficiency of genetic algorithm (GA), particle swarm optimization (PSO), differential evolution (DE), and artificial bee colony (ABC) in~\citep{Jamil2022}. D\'{i}az-\'{A}lvarez et al. investigated the impact of population size on the energy consumption of genetic programming (GP)~\citep{Diaz2022}. Fern\'{a}ndez de Vega et al. focused on parameters, including population size, affecting the energy efficiency of GAs~\citep{Fernandez2020}. Merelo-Guerv\'{o}s et al. studied EAs under different JavaScript interpreters and identified the most energy-efficient option, with energy scaling well with problem size~\citep{MeleloGuervos2024}. Cotta et al. demonstrated that energy consumption in batch-mode EA execution could be reduced by optimizing inter-run intervals~\citep{Cotta2024}.

While these studies provide insights into the general concept of incorporating energy into traditional EAs, they have not examined the interplay between surrogate modeling and energy consumption, a gap that this study aims to address.

\subsection{Surrogate Models and Their Efficiency/Accuracy}

Surrogate-assisted EAs (SAEAs) are widely adopted to reduce computational costs in expensive optimization problems, where objective evaluations are time-consuming~\citep{HE2023119495}. By approximating expensive objective functions, surrogates help reduce the number of actual evaluations.

Many studies on SAEAs focus on reducing computational time or the number of evaluations. For instance, Cui et al. evaluated algorithms under tight evaluation budgets~\citep{Cui2022SAEO}. Wei et al. proposed eToSA-DE, a two-stage surrogate-assisted DE for constraint-intensive problems, achieving reduced runtime despite training overheads~\citep{Wei2023eToSA-DE}. Zhang et al. demonstrated that a surrogate-assisted optimizer can reduce the time required to optimize energy cost and penalties in a regenerative Methanol cycle by 99\% compared to physics-based models~\citep{ZHANG2024}.

However, few studies explore the energy efficiency of surrogate-assisted approaches. One instance is \citep{Li2022PHEV}, which proposed SSPEA for multi-objective optimization in plug-in hybrid vehicle energy management. Their method reduced energy during the R\&D phase by over 44.6\%, showing that surrogate integration can reduce development costs and energy usage.

Despite these examples, the energy behavior of SAEAs remains poorly understood, particularly how surrogate training strategies (e.g., pre-training vs. retraining) impact energy efficiency during optimization. Our study addresses this gap by providing a comparative analysis of surrogate-assisted PSO and GA families under varying surrogate training strategies.

\section{Traffic Light Scheduling: A Costly Real-World Problem}\label{sec:problem}
Our focus in this work is on algorithms and surrogates, though they are obviously motivated by the need to solve computationally costly problems. We have then selected a difficult, useful task consisting of optimizing the network of traffic lights to provide sustainable mobility in real cities. Traffic lights indeed play a crucial role in managing traffic flow in cities. They are placed at intersections, pedestrian crossings, and other locations to control traffic and avoid accidents. At each intersection, the traffic lights are synchronized to work in a valid sequence of phases. Each phase includes a combination of colored lights and allows vehicles to use the roadway for a certain amount of time. The assignment of the duration for each phase in the sequence of all intersections in an urban area is called a traffic light plan. Finding an optimal traffic light plan is essential to minimize the number of stops at red lights, thereby reducing travel time for vehicles.

The task at hand involves optimizing multiple objectives for a specific area within a particular time frame. These objectives may include maximizing the number of vehicles that reach their destination within a given period, increasing the average speed of all vehicles in the area, reducing the waiting time of vehicles at intersections, and minimizing the length of waiting queues. To represent this problem mathematically, we adopt the approach used by \citep{GARCIANIETO2012274}. This is a multi-objective problem that aims to maximize the number of vehicles arriving at their destination ($NV_D$) and minimize the number of vehicles that do not reach their destination ($NV_{ND}$) during the simulation time ($T_S$). Additionally, the problem seeks to minimize the total travel time ($TT_v$) of all vehicles from the start of the simulation to their destination and the total waiting time of all vehicles ($TT_{EP}$). Furthermore, we maximize the ratio $P$ of green to red durations in each phase state of all intersections using the following equation:
\begin{equation}
P =\sum_{i=0}^{inter}\sum_{j=0}^{fs}d_{i,j}\frac{g_{i,j}}{r_{i,j}},
\label{eq: prob1}
\end{equation}

Here, $inter$ represents the number of intersections, $fs$ represents the number of phases in each intersection, and $g_{i,j}$ and $r_{i,j}$ denote the number of green and red signal colors at intersection $i$ and phase state $j$. $d_{i,j}$ represents the duration of these signals. Lastly, all objectives are combined into a single objective as:
\begin{equation}\label{eq:prob2}
F= \frac{TT_{v} + TT_{EP} + NV_{ND} \times T_{S}}{NV_{D}^{2}+P},
 \end{equation}

The end goal is to minimize Eq.~\eqref{eq:prob2}. A possible solution to this problem is a vector of positive integers where each integer represents the duration of each traffic light phase at each intersection in seconds. To evaluate the effectiveness of each potential solution, we need to analyze its impact on the traffic flow of the city under consideration. To achieve this, we use SUMO (Simulation of Urban MObility) to simulate the movement of vehicles in detail and collect the necessary data for calculating the quality of each potential solution according to Eq.~\eqref{eq:prob2}. While the results of the simulator are highly accurate and closely reflect reality, the process is time-consuming, taking several seconds to tens of minutes for each potential solution, depending on the size of the simulated urban area. This cost and computationally intensive process clearly highlights the need to explore ways to speed it up, such as using surrogate systems.

\subsection{Problem Instances}\label{subsec:problem_instances}
The experimental evaluation considers three real-world traffic light scheduling instances from \Malaga, Stockholm, and Paris, each characterized by different network sizes and traffic conditions (see Table~\ref{tab:problem_instances}). The \Malaga instance includes 56 intersections and 190 phases, managing 1,200 vehicles over a simulation period of 2,200 seconds. The Stockholm instance, representing a more complex urban environment, involves 75 intersections and 370 phases, with 1,400 vehicles and a simulation time of 4,000 seconds. Finally, the Paris instance, with 70 intersections and 378 phases, also models 1,200 vehicles, but over a longer simulation time of 3,400 seconds. These settings are based on several previous studies~\citep{VILLAGRA2020101085,Segredo2019} that addressed the same problem instances. These instances enable a broader analysis of the impact of surrogate-assisted optimization across different traffic densities and scenarios. 

\begin{table}[tb]
\centering
\caption{Information about the problem instances}
\label{tab:problem_instances} 
\vspace{-0.5em}
\scriptsize{
\begin{tabular}{lrrr}
\toprule
     &  \Malaga  & Stockholm &Paris\\
 \midrule
Total number of intersections& 56 &75&70\\
Total number of phases& 190 &370&378\\
Total number of vehicles&   1,200&  1,400&1,200\\
Simulation time (\si{\second})& 2,200   &4,000&3,400\\
\bottomrule
\end{tabular}
}
\vspace{-2em}
\end{table}

\section{Surrogate-Assisted Search Algorithms} \label{sec:surrogate_methods}
This section describes the surrogate-assisted search algorithms evaluated in this study, with a focus on their integration into PSO and GA. In both frameworks, NN surrogate models are incorporated to reduce the computational cost of repeated evaluations of the actual objective function. Two surrogate utilization strategies are considered: pre-training before optimization, and iterative retraining during the optimization process.

\subsection{Surrogate-Assisted Particle Swarm Optimization} \label{subsec: SAPSO}
The PSO algorithm is based on the PSO formulation originally proposed in the work~\citep{GARCIANIETO2012274} for the traffic light scheduling problem. Three configurations are considered: the baseline PSO without a surrogate model, SAPSO-p with a pre-trained surrogate, and SAPSO-r with a retraining-based surrogate.

Initially, $N$ candidate solutions are randomly generated as the initial swarm $P_0$, and evaluated using the actual objective function $f$.
The personal and global best positions are initialized, and the initial dataset $D$ is created to store evaluated solutions. Surrogate model training and retraining flags are initialized.

Each generation, particle velocities and positions are updated using the following standard PSO equations:
\begin{align}
    \bm{v}_{g+1}^i&\leftarrow w\bm{v}_g^i + \phi_1U(0, 1)(\bm{p}_g^i-\bm{x}_g^i)+\phi_2 U(0, 1)(\bm{b}_g-\bm{x}_g^i)
    \label{eq:update_velocity}\\
    \bm{x}_{g+1}^i&\leftarrow \bm{x}_{g}^i + \bm{v}_{g+1}^i
    \label{eq:update_position}
\end{align}
where $\bm{p}_g^i$ and $\bm{b}_g$ represent the personal and global best solutions, respectively. The inertia weight $w$, and acceleration coefficients $\phi_1$ and $\phi_2$ govern the balance between exploration and exploitation. $U(0, 1)$ corresponds to a uniform random number ranging from 0 to 1.

The velocity update is further modified to accommodate integer variables using stochastic rounding as follows:
\begin{equation}
    v_{g+1}^i=\begin{cases}
        \lfloor v_{g+1/2}^i\rfloor&\text{if}\:U(0, 1)^i\le \lambda\\
        \lceil v_{g+1/2}^i\rceil&\text{otherwise}
    \end{cases}
    \label{eq:velocity_calculation}
\end{equation}
where $v_{g+1/2}^i$ is the intermediate velocity value from Eq.~\eqref{eq:update_velocity}, while $\lfloor\cdot\rfloor$ and $\lceil\cdot\rceil$ indicate the floor and ceiling functions, respectively. The parameter $\lambda$ determines the probability of performing the floor or ceiling functions.

In addition, the inertia weight $w$ is linearly decreased as:
\begin{equation}
    w\leftarrow w_{max}-\frac{(w_{max}-w_{min})\times g}{g_{total}},
    \label{eq:weight_schedule}
\end{equation}
where $w_{max}$ and $w_{min}$ are the maximum (initial) and minimum (final) weights, while $g_{total}$ is the maximum generation count. 

SAPSO integrates an NN surrogate model $\hat{f}$ to reduce the number of expensive evaluations. When the dataset size $|D|$ exceeds a predefined threshold $N_t$, and training has not yet been completed (or retraining is active), the surrogate is trained. The use of the surrogate differs across the configurations:
\begin{description}
    \setlength\parskip{-1em}%
    \setlength\itemsep{0em}%
    \item[SAPSO-p:] The surrogate is trained once when $|D|\ge N_t$, and thereafter used exclusively to predict objective values.  The optimization relies entirely on $\hat{f}$, and a final evaluation is performed only on the best-predicted solution at the end.
    \setlength\parskip{0em}%
    \item[SAPSO-r:] In addition to the procedure of SAPSO-p, this configuration selects the top $N_r$ particles from each generation based on surrogate predictions and evaluates them using the actual objective function. These evaluated particles are added to $D$, and the surrogate model is retrained.
\end{description}
\vspace{-1em}

This framework enables comparison between fixed and adaptive surrogate strategies under the same PSO search structure.

\subsection{Surrogate-Assisted Genetic Algorithm}\label{subsec: SAGA}

This section describes the surrogate-assisted GA (SAGA), which extends a standard GA used in~\citep{Segredo2019} by integrating an NN surrogate model. The framework mirrors the structure of SAPSO described in Section~\ref{subsec: SAPSO}, but adapts the evolutionary operators in GA. 
Three configurations are considered: the baseline GA without a surrogate model, SAGA-p with a pre-trained surrogate, and SAGA-r with a retraining-based surrogate. 

The optimization begins with an initial population $P_0$ of $N$ solutions, which are evaluated using the actual objective function $f$. The dataset $D$ is initialized to store evaluated solutions, and flags for training and retraining are set accordingly.

At each generation, the offspring population $Q_g$ is initialized, and $N/2$ pairs of parents are selected from the current population $P_g$ based on a predefined selection strategy. For each pair, crossover and mutation operations are applied to produce two offspring, $o_1^\prime$ and $o_2^\prime$. Their objective values are either predicted by the surrogate $\hat{f}$ or evaluated using the actual function $f$, depending on whether the surrogate has been trained.
\begin{description}
    \setlength\parskip{-1em}%
    \setlength\itemsep{0em}%
\item[SAGA-p:] Once the size of the dataset $D$ reaches the threshold size $N_t$, a surrogate model $\hat{f}$ is trained and fixed for the remainder of the search. All subsequent offspring are evaluated using $\hat{f}$, and only the final best solution is evaluated with the true function.
    \setlength\parskip{0em}%
\item[SAGA-r:] In addition to the procedure of SAGA-p, a subset of $N_r$ offspring with the highest surrogate-predicted objective is selected and evaluated with the actual function. The resulting data are used to retrain the surrogate model, allowing it to dynamically adapt to the evolving population.
\end{description}
\vspace{-1em}

Following evaluation, the offspring are merged into $Q_g$, and the total number of function evaluations $FE$ is updated. The next generation population $P_{g+1}$ is formed by selecting the top $N$ individuals from the union of $P_g$ and $Q_g$, and the generation counter $g$ is incremented.

This design enables consistent comparison of fixed versus adaptive surrogate strategies under the evolutionary dynamics of GA, complementing the analysis performed with SAPSO.

\section{Experimental Setup}\label{sec:experimental_setup}
This section describes the system configuration and experimental environment used to evaluate the performance and energy profile of the surrogate-assisted optimization algorithms.
\subsection{System Specification}
All experiments were conducted on a computer running Ubuntu 22.04 with an Intel(R) Xeon(R) CPU E5-1650 v2 operating at \SI{3.50}{\giga\hertz} and having \SI{16}{\giga\byte} of DRAM memory.
The algorithms were implemented in Python 3.10.12. NN surrogates were developed using Tensorflow 2.17.0\footnote{\url{https://www.tensorflow.org/}} and Keras 3.5\footnote{\url{https://keras.io/}}. 
\subsection{Parameter Settings}
\begin{table}[tb]
\centering
\caption{Parameter settings of the NN surrogate model}
\label{tb:param_ann}
\vspace{-0.5em}
\scalebox{0.7}{
\begin{tabular}{lccc}
\toprule
&\multicolumn{3}{c}{Value}\\
\cmidrule(lr){2-4}
Parameter&\Malaga&Stockholm&Paris\\
\midrule
Number of input&190&370&378\\
Number of hidden layers&2&2&2\\
Number of hidden neurons&285--190&555--370&567--378\\
Activation function&\multicolumn{3}{c}{ReLU}\\
Optimizer&\multicolumn{3}{c}{Adam}\\ 
Number of training epochs& \multicolumn{3}{c}{100}\\
Batch size&\multicolumn{3}{c}{32}\\
Learning rate&\multicolumn{3}{c}{$1.0\times 10^{-4}$}\\
\bottomrule
\end{tabular}
}
\vspace{-1.5em}
\end{table}
Table~\ref{tb:param_ann} summarizes the NN architecture used for the surrogate models. The number of input neurons corresponds to the dimensionality of the traffic light scheduling problem: 190 for \Malaga, 370 for Stockholm, and 378 for Paris. 

Each NN consists of two hidden layers. The first hidden layer contains 1.5 times the number of input neurons, and the second hidden layer matches the number of input neurons. For instance, the \Malaga instance has hidden layers of 285 and 190 neurons, respectively. Similarly, Stockholm and Paris have hidden layers of 555--370 and 567--378 neurons, respectively.

The hidden layers use the ReLU activation function. The NN is trained with the Adam optimizer for 100 epochs, using a batch size of 32 and a learning rate of $1.0\times 10^{-4}$. The training loss function is a mean squared error (MSE).

\subsection{Measurement of Energy and Memory Usage}
To measure energy consumption, we used the Python package pyRAPL\footnote{\url{https://pyrapl.readthedocs.io/en/latest/}}, which leverages Intel's Running Average Power Limit (RAPL) technology~\citep{David2010}. Intel RAPL is a reliable standard for energy analysis~\citep{Abdelhafez2019,Diaz2022,Ferro2023,anthony2020carbontracker}. The pyRAPL enables the separate measurement of the energy of the CPU and DRAM.

For memory usage analysis, we employed Python's built-in \texttt{tracemalloc} module, which tracks memory allocations over time, providing detailed insights into the algorithm's memory behavior. In addition, since \texttt{tracemalloc} does not capture the memory usage of subprocesses, the memory usage of the SUMO simulator is measured using the \texttt{time} command.

\section{On the Cost of Evaluating a Tentative Solution}\label{sec:solution_evaluation}
This section examines the computational characteristics of solution evaluation, focusing on energy consumption, execution time, and memory usage. Section~\ref{sec:Eval_procedure} describes the measurement procedure, and Section~\ref{sec:Eval_result} presents a statistical analysis of the recorded metrics. 

\subsection{Statistical Validation of the Computed Costs}\label{sec:Eval_procedure}
To measure the computational cost of actual solution evaluation, 50,000 randomly generated solutions were evaluated using the SUMO simulator while measuring CPU and DRAM energy consumption, execution time, and memory usage.

To model and analyze the distribution of computational costs, a log-normal distribution is fitted to the measured values. This distribution was selected over others like normal or beta not only for its ability to model asymmetric and long-tailed data effectively (see Fig.~\ref{fig:solution_evaluacion_distributions}), but also for its interpretability and parsimony, as it requires only two parameters.
The probability density function of the log-normal distribution is given by:
\begin{equation}
f(x) = \frac{1}{x \sigma \sqrt{2\pi}} \exp\left(-\frac{(\ln x - \mu)^2}{2\sigma^2}\right),\quad 0 < x < \infty,
\label{eq:lognormal}
\end{equation}
where $\mu$ and $\sigma$ denote the mean and standard deviation of the natural logarithm of the observations. The mean $E(X)$ and standard deviation $SD(X)$ on the original scale are given by:
\begin{align}
E(X) &= \exp\left(\mu + \frac{\sigma^2}{2}\right)\label{eq:log_mean} \\
SD(X) &= \sqrt{(\exp(\sigma^2) - 1) \exp(2\mu + \sigma^2)}\label{eq:log_std}.
\end{align}

To evaluate the model's goodness-of-fit, we use the Normalized Mean Squared Error (NMSE):
\begin{equation}
\text{NMSE} =
\frac{\sum_{i=1}^{B} (f_{\text{obs}}(x_i) - f_{\text{model}}(x_i))^2}
{\sum_{i=1}^{B} f_{\text{obs}}(x_i)^2},
\label{eq:mse}
\end{equation}
where $f_{\text{model}}(x)$ is the fitted log-normal distribution, $f_{\text{obs}}(x)$ is the empirical density, and $B = 50$ denotes the number of bins. Lower NMSE values indicate better fits, while values close to or exceeding 1 suggest poor fit quality.

\subsection{Analysis of Time, Memory, and Energy Consumption}\label{sec:Eval_result}
\begin{table*}[tb]
\tabcolsep = 5.4pt
\centering
\caption{Statistical characterisation of solution evaluation. Mean and standard deviation (Eqs.~\eqref{eq:log_mean} and \eqref{eq:log_std}), $\mu$ and $\sigma$ (Eq.~\eqref{eq:lognormal}), and NMSE (Eq.~\eqref{eq:mse}).}
\label{tb:sol_eval_results}
\vspace{-0.5em}
\scriptsize{
\begin{tabular}{lrrrrrrrrrrrrrrr}
\toprule
&\multicolumn{5}{c}{\Malaga}&\multicolumn{5}{c}{Stockholm}&\multicolumn{5}{c}{Paris}\\
\cmidrule(lr){2-6}\cmidrule(lr){7-11}\cmidrule(lr){12-16}
&$E$&$SD$&$\mu$&$\sigma$&NMSE&$E$&$SD$&$\mu$&$\sigma$&NMSE&$E$&$SD$&$\mu$&$\sigma$&NMSE\\
\midrule
CPU (\si{\joule}) & \textbf{216.38} & 29.72 & 5.37 & 0.14 & $6.46\times10^{-3}$ & 543.35 & 85.75 & 6.29 & 0.16 & $6.52\times10^{-3}$ & 341.27 & 39.05 & 5.83 & 0.11 & $9.74\times10^{-3}$ \\
DRAM (\si{\joule}) & \textbf{12.57} & 1.80 & 2.52 & 0.14 & $1.86\times10^{-3}$ & 27.91 & 5.16 & 3.31 & 0.18 & $1.10\times10^{-3}$ & 18.96 & 2.50 & 2.93 & 0.13 & $1.01\times10^{-3}$ \\
Total Energy (\si{\joule}) & \textbf{228.95} & 31.35 & 5.42 & 0.14 & $6.34\times10^{-3}$ & 571.25 & 89.98 & 6.34 & 0.16 & $6.14\times10^{-3}$ & 360.22 & 41.14 & 5.88 & 0.11 & $9.45\times10^{-3}$ \\
Time (\si{\second}) & \textbf{4.26} & 0.58 & 1.44 & 0.14 & $6.28\times10^{-3}$ & 10.68 & 1.68 & 2.36 & 0.16 & $6.39\times10^{-3}$ & 6.72 & 0.76 & 1.90 & 0.11 & $9.65\times10^{-3}$ \\
Memory usage (\si{\mega\byte}) & 44.81 & 0.19 & 3.80 & 0.00 & $2.97\times10^{-3}$ & 44.81 & 0.19 & 3.80 & 0.00 & $4.83\times10^{-3}$ & \textbf
{44.81} & 0.19 & 3.80 & 0.00 & $5.10\times10^{-3}$ \\
\bottomrule
\end{tabular}
}
\vspace{-1em}
\end{table*}
\begin{figure*}[!tb]
    \centering
    \begin{minipage}{0.5\textwidth} 
        \includegraphics[width=\linewidth]{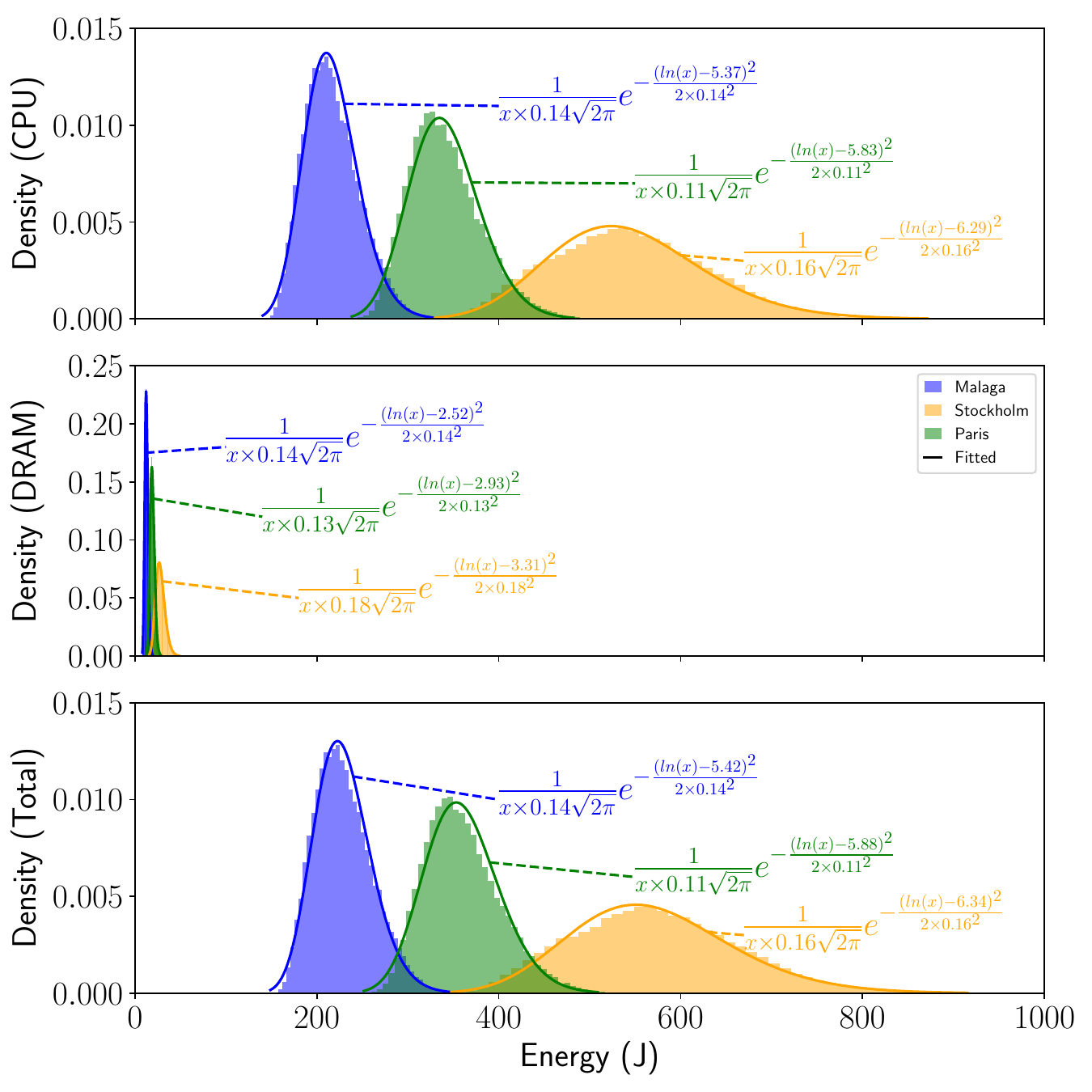}
       \subcaption{Energy consumption of CPU and DRAM, and their total}
    \end{minipage}
    \hfill
    \begin{minipage}{0.46\textwidth} 
    \begin{minipage}{\textwidth} 
        \includegraphics[width=\linewidth]{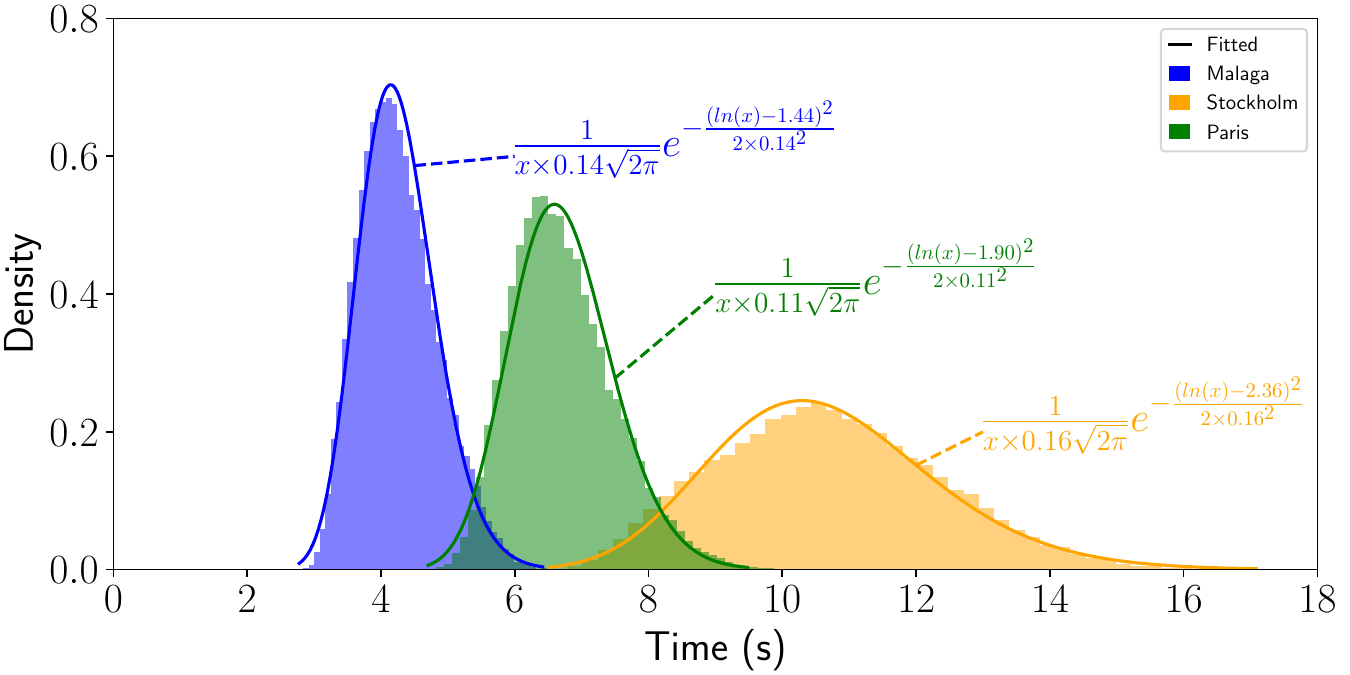}
       \subcaption{Execution time}
    \end{minipage}
    \begin{minipage}{\textwidth} 
        \includegraphics[width=\linewidth]{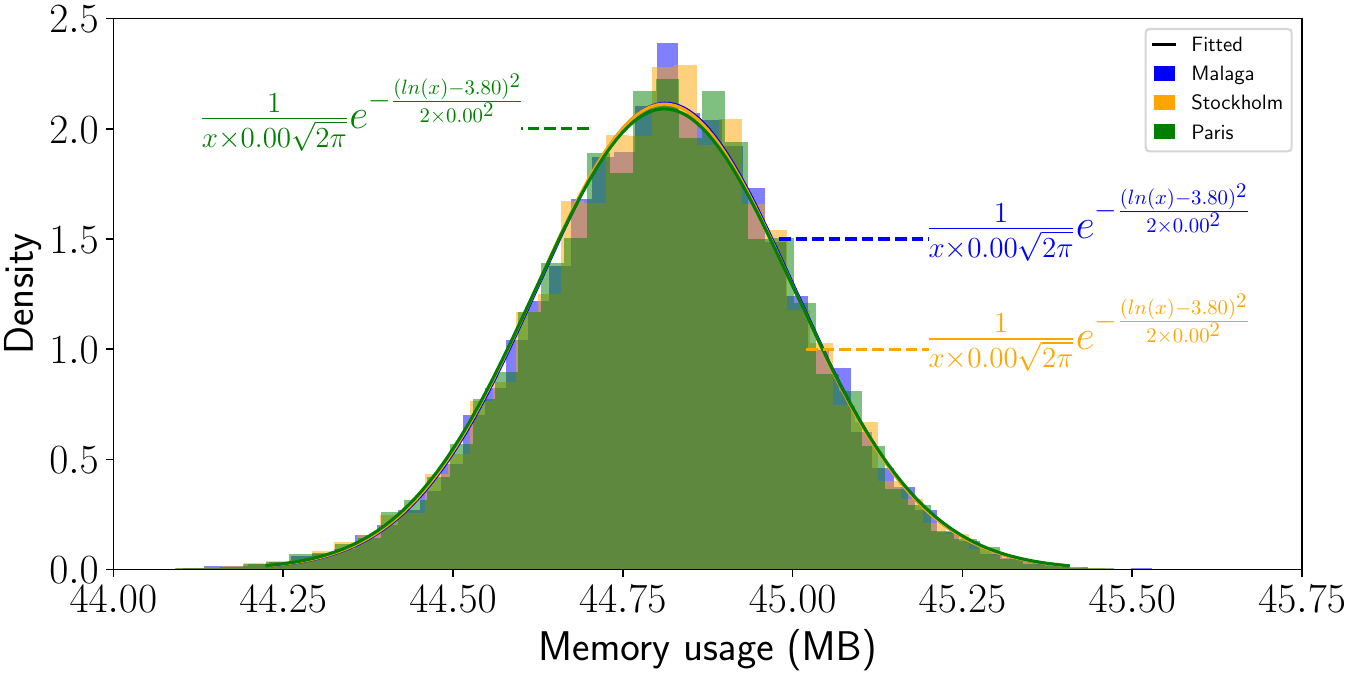}
       \subcaption{Memory usage}
    \end{minipage}
    \end{minipage}
\vspace{-0.5em}
\caption{Distribution of the energy consumption, execution time, and memory usage for solution evaluations\label{fig:solution_evaluacion_distributions}}
\vspace{-1.5em}
\end{figure*}
Table~\ref{tb:sol_eval_results} summarizes the statistical characteristics of CPU and DRAM energy consumption, execution time, and memory usage for solution evaluations across the \Malaga, Stockholm, and Paris instances. The minimum values among the instances are highlighted in bold in this table. Fig.~\ref{fig:solution_evaluacion_distributions} presents the empirical distributions and corresponding log-normal fits for each metric. 

In the \Malaga instance, computational cost is the lowest, with average energy consumption of \SI{228.95}{\joule} and execution time of \SI{4.26}{\second}. CPU and DRAM energy consumption follow the same trend. Memory usage is \SI{44.81}{\mega\byte} with a small variation.

Stockholm exhibits the highest computational cost. Energy and time reach \SI{571.25}{\joule} and \SI{10.68}{\second}, respectively, with CPU and DRAM consumption peaking accordingly. Memory usage is \SI{44.81}{\mega\byte}, which is nearly identical to that of \Malaga, indicating no correlation with increased computational effort.

In Paris, the computational cost is intermediate, with an energy consumption of \SI{360.22}{\joule} and an execution time of \SI{6.72}{\second}. Memory usage remains unchanged at \SI{44.81}{\mega\byte}, indicating minimal influence from problem dimensionality.

All metrics are well approximated by log-normal distributions, as indicated by low NMSE values. Fig.~\ref{fig:solution_evaluacion_distributions} confirms that energy consumption and execution time increase in the order of \Malaga, Paris, and Stockholm, while memory usage distributions are nearly identical across instances. This is supported by the Kruskal–Wallis test, which found no significant differences in memory usage ($p\text{-value} = 0.666$).

\begin{figure}[!tb]
\centering
\begin{minipage}[t]{0.48\columnwidth}
   \includegraphics[width=\textwidth]{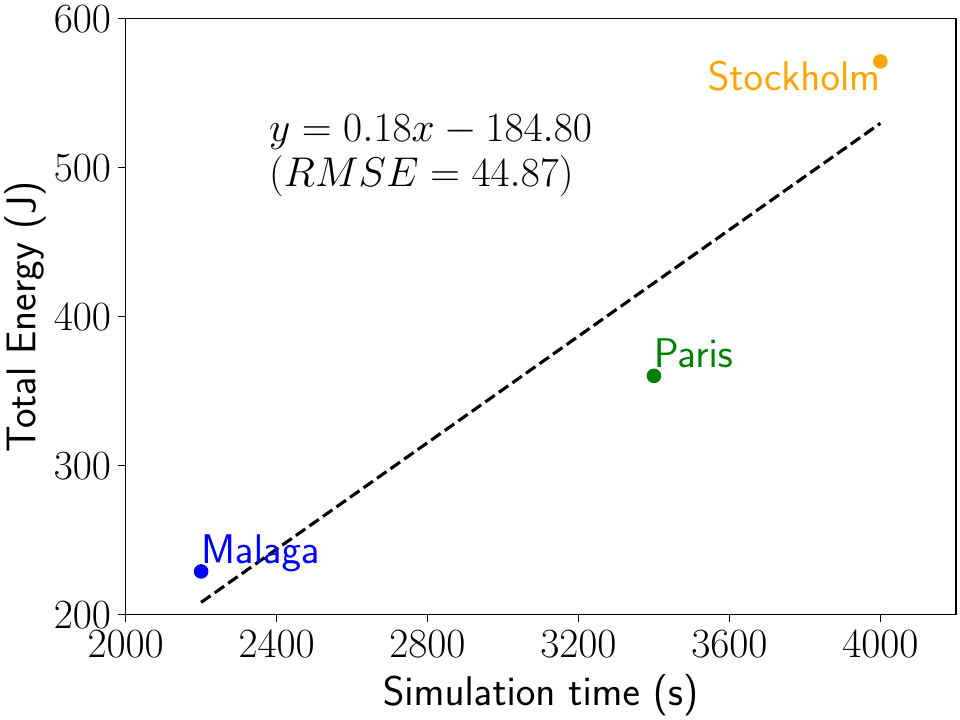} 
   \subcaption{Total energy consumption\label{fig:fitting_instance_energy}}
\end{minipage}
\hfill
\begin{minipage}[t]{0.48\columnwidth}
   \includegraphics[width=\textwidth]{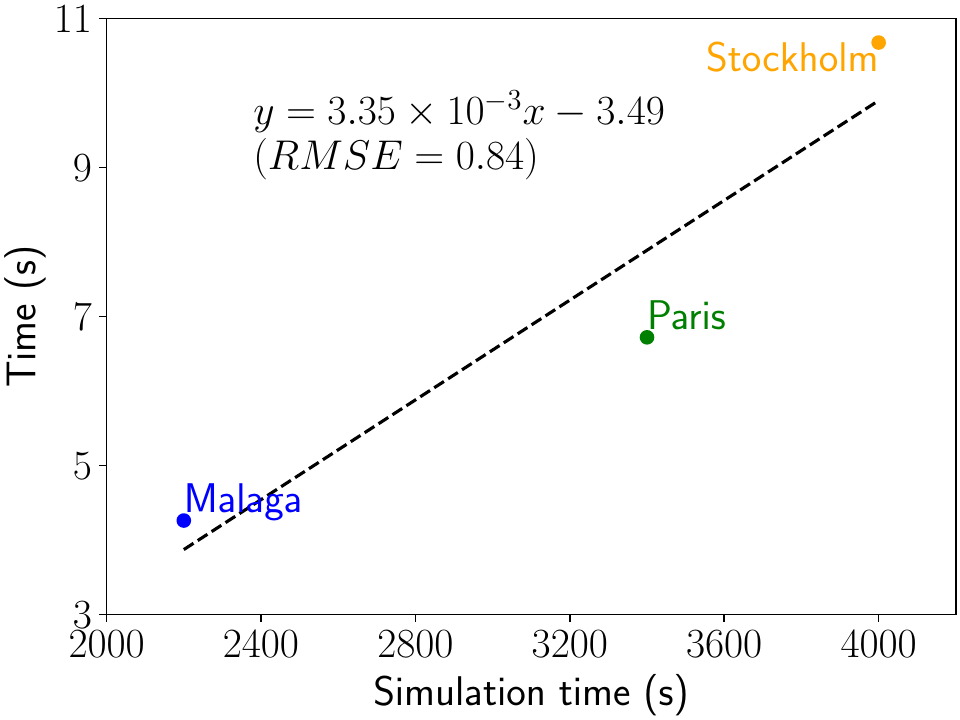} 
   \subcaption{Execution time\label{fig:fitting_instance_time}}
\end{minipage}
\vspace{-0.5em}
\caption{Linear regression for mean $E$ of energy and time by simulation time}
\label{fig:fitting_instance_energy_time}
\vspace{-1.5em}
\end{figure}
We calculated the Pearson correlation between the computational metrics (energy and time) and the instance-specific parameters in Table~\ref{tab:problem_instances}, and found a strong positive correlation ($\rho = 0.95$) between simulation time and both metrics. Fig.~\ref{fig:fitting_instance_energy_time} visualizes this relationship using linear regression models fitted to the average values per instance. The root mean squared error (RMSE) is shown within each plot. In this figure, the horizontal axis represents simulation time, while the vertical axes correspond to total energy consumption and execution time, respectively. The regression lines and RMSE values together confirm that both metrics increase consistently with simulation time across the three instances.

The results show consistent increases in both metrics with simulation time, with estimated growth rates:
\begin{itemize}
    \setlength\parskip{-1em}%
    \setlength\itemsep{0em}%
\item Energy: \SI{0.18}{\joule} increase per simulation second.
    \setlength\parskip{0em}%
\item Time: \SI{3.35}{\milli\second} increase per simulation second.
\end{itemize}
\vspace{-1em}

These findings confirm that the simulation time is the dominant factor affecting energy and execution time. These insights highlight the significance of instance characteristics in determining computational resource demands, a consideration particularly crucial in large-scale simulations where both accuracy and efficiency must be balanced.

\section{Training and Using ML Surrogates}\label{sec:nn_training}
This section analyzes the computational performance of NN training and utilization. Section~\ref{sec:NN_eval_procedure} describes the measurement procedure, followed by an analysis of training and prediction performance in Sections~\ref{sec:NN_training} and \ref{sec:NN_use}, respectively. Section~\ref{sec:NN_structure} examines the internal structure of the NN, and Section~\ref{sec:NN_energy_benefits} discusses the long-term energy benefits of large training datasets.

\subsection{Grid Optimization of the Dataset Size}\label{sec:NN_eval_procedure}
In this experiment, 100 NNs were trained using distinct training dataset sizes: $N_t = \{128, 256, 512, 1024, 2048, 4096, 8192\}$. The training data were selected uniformly at random from the archive obtained in Section~\ref{sec:Eval_procedure}, without replacement. For testing, 100 additional solutions were uniformly sampled from the remaining archive entities, ensuring no overlap with the training data. Each trained NN was then used to predict the objective values of the corresponding test solutions. This experiment aims to determine the optimal balance between training effort and prediction accuracy of the surrogate models.

The next subsection first analyzes the computational performance for the NN training.

\subsection{NN Training}\label{sec:NN_training}
\begin{table*}[!tb]
\centering
\caption{Total energy (\si{\joule}) for NN training. Mean and standard deviation calculated by Eqs.~\eqref{eq:log_mean} and \eqref{eq:log_std}, the parameters $\mu$ and $\sigma$ in Eq.~\eqref{eq:lognormal}, and NMSE.}
\label{tb:NN_train_total}
\vspace{-0.5em}
\scriptsize{
\begin{tabular}{lrrrrrrrrrrrrrrr}
\toprule
&\multicolumn{5}{c}{M\'{a}laga}&\multicolumn{5}{c}{Stockholm}&\multicolumn{5}{c}{Paris}\\
\cmidrule(lr){2-6}\cmidrule(lr){7-11}\cmidrule(lr){12-16}
\# dataset&$E$&$SD$&$\mu$&$\sigma$&NMSE&$E$&$SD$&$\mu$&$\sigma$&NMSE&$E$&$SD$&$\mu$&$\sigma$&NMSE\\
\midrule
128 & \textbf{217.79} & 1.02 & 5.38 & 0.00 & $ 1.96 \times 10^{-1}$ & \textbf{264.40} & 0.99 & 5.58 & 0.00 & $ 2.57 \times 10^{-1}$ & \textbf{265.10} & 2.01 & 5.58 & 0.01 & $ 2.27 \times 10^{-1}$ \\
256 & 255.35 & 5.90 & 5.54 & 0.02 & $ 6.68 \times 10^{-1}$ & 329.25 & 3.65 & 5.80 & 0.01 & $ 2.41 \times 10^{-1}$ & 334.03 & 3.48 & 5.81 & 0.01 & $ 2.36 \times 10^{-1}$ \\
512 & 302.47 & 2.51 & 5.71 & 0.01 & $ 2.36 \times 10^{-1}$ & 493.70 & 3.37 & 6.20 & 0.01 & $ 1.83 \times 10^{-1}$ & 501.60 & 3.29 & 6.22 & 0.01 & $ 2.44 \times 10^{-1}$ \\
1024 & 445.26 & 3.68 & 6.10 & 0.01 & $ 2.34 \times 10^{-1}$ & 828.99 & 7.86 & 6.72 & 0.01 & $ 3.29 \times 10^{-1}$ & 850.19 & 6.86 & 6.75 & 0.01 & $ 1.78 \times 10^{-1}$ \\
2048 & 735.38 & 5.22 & 6.60 & 0.01 & $ 1.38 \times 10^{-1}$ & 1497.63 & 14.47 & 7.31 & 0.01 & $ 2.97 \times 10^{-1}$ & 1539.46 & 12.57 & 7.34 & 0.01 & $ 2.49 \times 10^{-1}$ \\
4096 & 1333.51 & 13.80 & 7.20 & 0.01 & $ 3.64 \times 10^{-1}$ & 2867.36 & 27.95 & 7.96 & 0.01 & $ 2.73 \times 10^{-1}$ & 2922.56 & 29.85 & 7.98 & 0.01 & $ 2.67 \times 10^{-1}$ \\
8192 & 2574.66 & 26.51 & 7.85 & 0.01 & $ 2.69 \times 10^{-1}$ & 5510.77 & 74.26 & 8.61 & 0.01 & $ 2.56 \times 10^{-1}$ & 5678.00 & 64.93 & 8.64 & 0.01 & $ 2.13 \times 10^{-1}$ \\
\midrule
$p$-value&\multicolumn{5}{l}{$<0.01$}&\multicolumn{5}{l}{$<0.01$}&\multicolumn{5}{l}{$<0.01$}\\
\bottomrule
\end{tabular}
}
\vspace{1em}
\centering
\caption{Execution time (\si{\second}) for NN training. Mean and standard deviation calculated by Eqs.~\eqref{eq:log_mean} and \eqref{eq:log_std}, the parameters $\mu$ and $\sigma$ in Eq.~\eqref{eq:lognormal}, and NMSE.}
\label{tb:NN_train_time}
\vspace{-0.5em}
\scriptsize{
\begin{tabular}{lrrrrrrrrrrrrrrr}
\toprule
&\multicolumn{5}{c}{M\'{a}laga}&\multicolumn{5}{c}{Stockholm}&\multicolumn{5}{c}{Paris}\\
\cmidrule(lr){2-6}\cmidrule(lr){7-11}\cmidrule(lr){12-16}
\# dataset&$E$&$SD$&$\mu$&$\sigma$&NMSE&$E$&$SD$&$\mu$&$\sigma$&NMSE&$E$&$SD$&$\mu$&$\sigma$&NMSE\\
\midrule
128 & \textbf{3.94} & 0.02 & 1.37 & 0.00 & $ 2.06 \times 10^{-1} $ & \textbf{4.51} & 0.02 & 1.51 & 0.00 & $ 1.74 \times 10^{-1} $ & \textbf{4.55} & 0.02 & 1.51 & 0.00 & $ 1.39 \times 10^{-1} $ \\
256 & 4.61 & 0.04 & 1.53 & 0.01 & $ 4.00 \times 10^{-1} $ & 5.97 & 0.04 & 1.79 & 0.01 & $ 1.86 \times 10^{-1} $ & 6.03 & 0.03 & 1.80 & 0.01 & $ 1.63 \times 10^{-1} $ \\
512 & 6.22 & 0.02 & 1.83 & 0.00 & $ 3.15 \times 10^{-1} $ & 8.65 & 0.05 & 2.16 & 0.01 & $ 2.29 \times 10^{-1} $ & 8.72 & 0.04 & 2.17 & 0.00 & $ 2.34 \times 10^{-1} $ \\
1024 & 9.41 & 0.06 & 2.24 & 0.01 & $ 1.74 \times 10^{-1} $ & 13.95 & 0.10 & 2.64 & 0.01 & $ 2.30 \times 10^{-1} $ & 14.16 & 0.09 & 2.65 & 0.01 & $ 1.56 \times 10^{-1} $ \\
2048 & 18.09 & 0.33 & 2.90 & 0.02 & $ 2.17 \times 10^{-1} $ & 24.34 & 0.19 & 3.19 & 0.01 & $ 2.11 \times 10^{-1} $ & 24.83 & 0.19 & 3.21 & 0.01 & $ 2.44 \times 10^{-1} $ \\
4096 & 41.09 & 0.79 & 3.72 & 0.02 & $ 2.22 \times 10^{-1} $ & 45.89 & 0.40 & 3.83 & 0.01 & $ 2.48 \times 10^{-1} $ & 46.37 & 0.43 & 3.84 & 0.01 & $ 2.70 \times 10^{-1} $ \\
8192 & 108.52 & 1.20 & 4.69 & 0.01 & $ 2.52 \times 10^{-1} $ & 87.43 & 1.10 & 4.47 & 0.01 & $ 2.65 \times 10^{-1} $ & 89.34 & 1.01 & 4.49 & 0.01 & $ 2.46 \times 10^{-1} $ \\
\midrule
$p$-value&\multicolumn{5}{l}{$<0.01$}&\multicolumn{5}{l}{$<0.01$}&\multicolumn{5}{l}{$<0.01$}\\
\bottomrule
\end{tabular}
}
\vspace{1em}
\centering
\caption{Memory usage (\si{\mega\byte}) for NN training. Mean and standard deviation calculated by Eqs.~\eqref{eq:log_mean} and \eqref{eq:log_std}, the parameters $\mu$ and $\sigma$ in Eq.~\eqref{eq:lognormal}, and NMSE.}
\label{tb:NN_train_memory}
\vspace{-0.5em}
\scriptsize{
\begin{tabular}{lrrrrrrrrrrrrrrr}
\toprule
&\multicolumn{5}{c}{M\'{a}laga}&\multicolumn{5}{c}{Stockholm}&\multicolumn{5}{c}{Paris}\\
\cmidrule(lr){2-6}\cmidrule(lr){7-11}\cmidrule(lr){12-16}
\# dataset&$E$&$SD$&$\mu$&$\sigma$&NMSE&$E$&$SD$&$\mu$&$\sigma$&NMSE&$E$&$SD$&$\mu$&$\sigma$&NMSE\\
\midrule
128 & \textbf{2.71} & 0.04 & 1.00 & 0.02 & $ 8.36 \times 10^{-1} $ & \textbf{3.09} & 0.08 & 1.13 & 0.02 & $ 8.95 \times 10^{-1} $ & \textbf{3.11} & 0.08 & 1.13 & 0.02 & $ 8.66 \times 10^{-1} $ \\
256 & 3.18 & 0.09 & 1.16 & 0.03 & $ 9.35 \times 10^{-1} $ & 3.94 & 0.17 & 1.37 & 0.04 & $ 9.25 \times 10^{-1} $ & 3.97 & 0.17 & 1.38 & 0.04 & $ 9.49 \times 10^{-1} $ \\
512 & 4.04 & 0.15 & 1.39 & 0.04 & $ 9.56 \times 10^{-1} $ & 5.79 & 0.36 & 1.75 & 0.06 & $ 9.72 \times 10^{-1} $ & 5.79 & 0.37 & 1.75 & 0.06 & $ 9.71 \times 10^{-1} $ \\
1024 & 5.63 & 0.30 & 1.73 & 0.05 & $ 9.48 \times 10^{-1} $ & 9.13 & 0.32 & 2.21 & 0.03 & $ 9.66 \times 10^{-1} $ & 9.49 & 0.37 & 2.25 & 0.04 & $ 9.75 \times 10^{-1} $ \\
2048 & 9.36 & 0.31 & 2.24 & 0.03 & $ 9.69 \times 10^{-1} $ & 17.63 & 0.00 & 2.87 & 0.00 & $ 7.89 \times 10^{-1} $ & 18.00 & 0.00 & 2.89 & 0.00 & $ 7.03 \times 10^{-1} $ \\
4096 & 18.11 & 0.03 & 2.90 & 0.00 & $ 8.71 \times 10^{-1} $ & 34.99 & 0.00 & 3.56 & 0.00 & $ 7.62 \times 10^{-1} $ & 35.74 & 0.00 & 3.58 & 0.00 & $ 7.34 \times 10^{-1} $ \\
8192 & 35.95 & 0.00 & 3.58 & 0.00 & $ 7.95 \times 10^{-1} $ & 69.71 & 0.02 & 4.24 & 0.00 & $ 8.40 \times 10^{-1} $ & 71.21 & 0.01 & 4.27 & 0.00 & $ 8.49 \times 10^{-1} $ \\
\midrule
$p$-value&\multicolumn{5}{l}{$<0.01$}&\multicolumn{5}{l}{$<0.01$}&\multicolumn{5}{l}{$<0.01$}\\
\bottomrule
\end{tabular}
}
\vspace{-2em}
\end{table*}
Tables~\ref{tb:NN_train_total}--\ref{tb:NN_train_memory} summarize the computational performance of NN training, including energy consumption, execution time, and memory usage. For each dataset size, the tables present the mean, standard deviation, and NMSE between the empirical distribution and the fitted log-normal distributions. Minimum values for each instance are highlighted in bold, and the bottom row provides the $p$-value of the Kruskal-Wallis test, which assesses whether significant differences exist among dataset sizes.

Table~\ref{tb:NN_train_total} shows that total energy consumption increases with dataset size for all instances, as expected due to the higher computational demand of training with larger datasets. The consistently low NMSE values indicate a good log-normal fit.

Table~\ref{tb:NN_train_time} presents similar trends for time, which also increases with dataset size due to the greater number of training iterations. NMSE values remain low, confirming a reasonable log-normal approximation, though minor deviations are present.

Table~\ref{tb:NN_train_memory} shows that memory usage remains nearly constant for each dataset size, primarily dictated by the NN structure and input dimensionality rather than the immediate computational load. However, memory usage still increases with larger datasets due to the greater storage requirements for training.

Across instances, energy consumption is consistently higher in Stockholm and Paris than in \Malaga, which is attributed to their higher input dimensionality and larger network size. Between Stockholm and Paris, energy is slightly lower in Paris, reflecting its marginally lower dimensionality. Execution time follows a similar trend: \Malaga is the fastest, followed by Paris, with Stockholm requiring the longest training time. An exception occurs for the dataset size of 8192, where \Malaga unexpectedly shows the longest training time, possibly due to overheads not directly related to dimensionality.

Memory usage is lowest for \Malaga, consistent with its smaller NN size. Stockholm and Paris exhibit slightly higher but comparable memory usage, suggesting that memory requirements are primarily determined by network architecture rather than dynamic computational factors.

The results confirm a reasonable trend. Specifically, as the dataset size increases, the energy and time required also increase because more iterations are needed to train the NN. Additionally, the increase in memory usage is justified by the need to store training data. While energy and time follow a log-normal distribution across dataset sizes, memory usage remains relatively constant within each dataset size. Furthermore, computational performance depends not only on dataset size but also on the problem dimensionality and NN architecture.

A direct comparison with solution evaluations reveals that, with a small dataset (size 128), NN training requires comparable or less energy. However, with a dataset size of 8192, NN training consumes approximately 11 times more energy in \Malaga, 9 times more in Stockholm, and 15 times more in Paris. Execution time shows a similar pattern, with a dataset size of 128, NN training and solution evaluations take a similar or shorter time, while with a dataset size of 8192, NN training takes up to 25 times, 8 times, and 14 times longer execution time, respectively. Memory usage shows a different trend. In \Malaga, even with a dataset size of 8192, NN training uses less memory than solution evaluations. In Stockholm and Paris, NN training requires less memory for a dataset size of 4096 or smaller, but about 1.5 times more for a dataset size of 8192. These findings highlight that, while NN training can reduce computational cost compared to solution evaluations for small datasets, larger datasets significantly increase training cost, particularly in energy consumption and execution time.

To approximate how dataset size and NN complexity affect computational performance, energy, time, and memory usage were modeled as functions of the number of network parameters ($n_p$) and dataset size ($n_t$). The initial model included all second-order terms and interactions of $n_p$ and $n_t$ as:
\begin{equation}
    y = a_1 n_p^2 + a_2 n_t^2 + a_3 n_p n_t+ a_4 n_p + a_5 n_t + a_6,
\end{equation}
where $y$ is the target metric, and $a_1$--$a_6$ are regression coefficients. Statistical analysis revealed that $n_p$ and $n_p^2$ were insignificant ($p\text{-value}> 0.05$). Thus, a simplified and interpretable model was derived by retaining only the significant predictors: $n_t^2$, $n_p n_t$, and $n_t$. The final model is expressed as:
\begin{equation}
    y = a_2^\prime n_t^2 + a_3^\prime n_p n_t + a_5^\prime n_t + a_6^\prime,
\end{equation}
with coefficients $a_2^\prime$, $a_3^\prime$, $a_5^\prime$, and $a_6^\prime$ re-estimated accordingly. 

\begin{figure*}[!tb]
    \centering
    \begin{minipage}[t]{0.32\textwidth}
    \includegraphics[width=\textwidth]{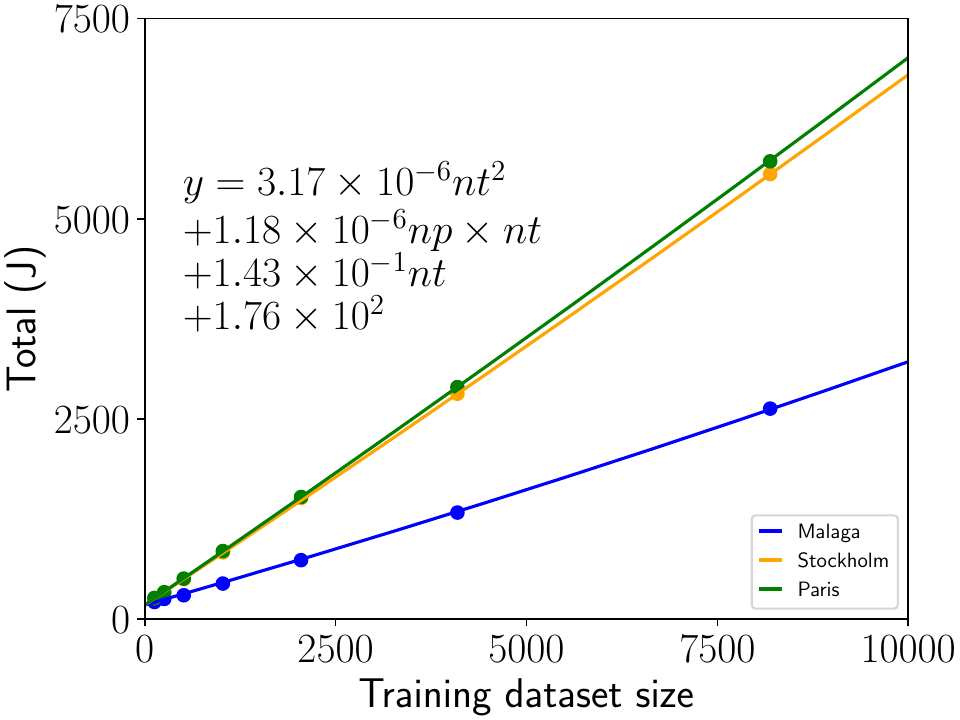}
    \subcaption{Total energy consumption}
    \label{fig:training_polyfit_energy}
    \end{minipage}
    \begin{minipage}[t]{0.32\textwidth}
    \includegraphics[width=\textwidth]{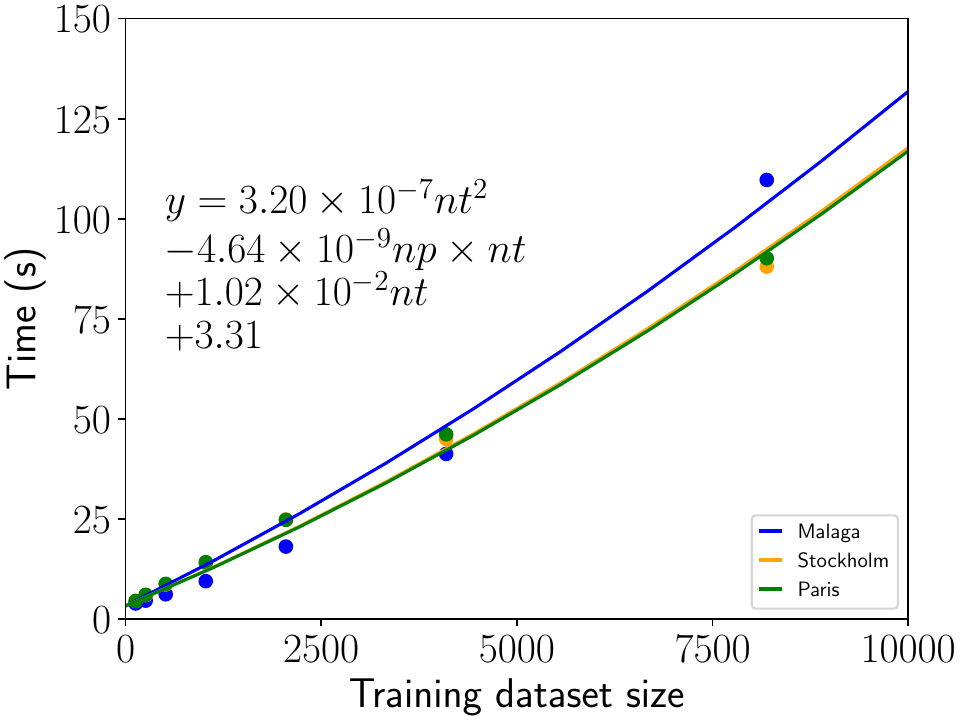}
    \subcaption{Execution time}
    \label{fig:training_polyfit_time}
    \end{minipage}
    \begin{minipage}[t]{0.32\textwidth}
    \includegraphics[width=\textwidth]{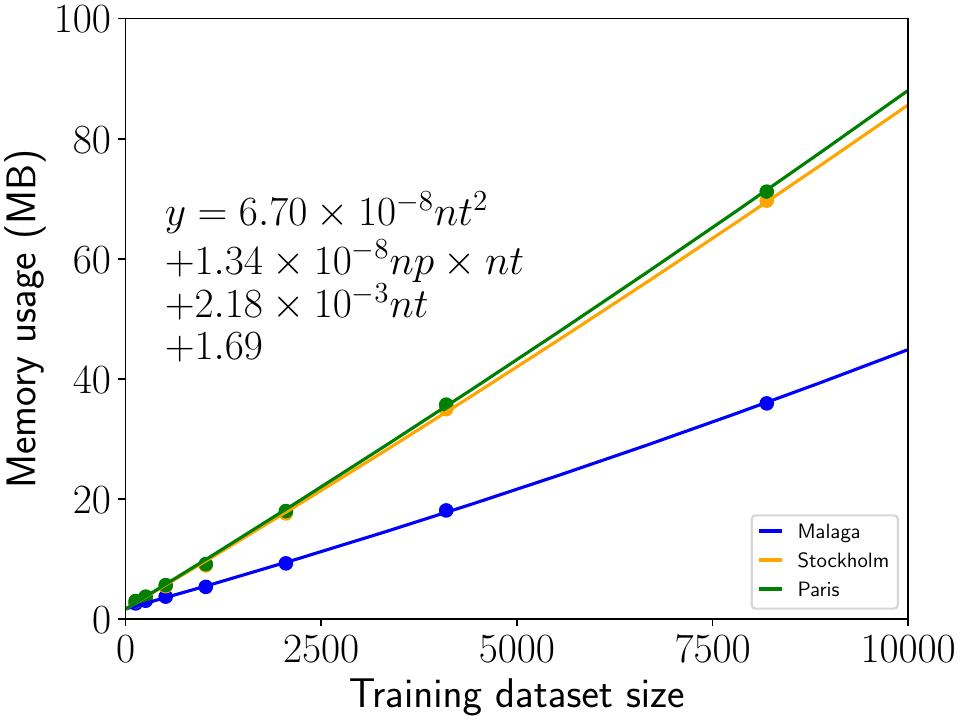}
    \subcaption{Memory usage}
    \label{fig:training_polyfit_memory}
    \end{minipage}
\vspace{-0.5em}
    \caption{Polynomial fitting with the number of parameters in NN ($np$) and training dataset size ($nt$)}
    \label{fig:training_polyfit}
    \vspace{-1.5em}
\end{figure*}
Fig.~\ref{fig:training_polyfit} illustrates the relationship between dataset size and mean total energy (Fig.~\ref{fig:training_polyfit_energy}), time (Fig.~\ref{fig:training_polyfit_time}), and memory usage (Fig.~\ref{fig:training_polyfit_memory}) across instances. The horizontal axis represents the dataset size, while the vertical axis corresponds to each measurement. Different colors indicate the different instances.

Fig.~\ref{fig:training_polyfit} confirms that the proposed approximation effectively models energy consumption, execution time, and memory usage as functions of dataset size and network complexity. The fitted curves align well with the observed values, demonstrating that the given formulation can reasonably capture the relationship. However, for execution time, a noticeable deviation is observed in the \Malaga instance when the dataset size is 8192, where the measured execution time exceeds the predicted value. This suggests that additional computational overheads associated with large dataset sizes may be present, which are not accounted for in the current model.

We now turn to the performance of NN utilization, where the trained models are used to predict solution evaluations.

\subsection{NN Utilization}\label{sec:NN_use}
Tables~\ref{tb:NN_use_total}--\ref{tb:NN_use_memory} summarize the computational performance metrics for NN utilization, including energy, time, and memory usage. These tables present the mean and standard deviation estimated from the observed data, along with NMSE values comparing the empirical distributions to the fitted log-normal models. The minimum values for each instance are highlighted in bold. Additionally, the bottom row in each table reports the $p$-value of the Kruskal-Wallis test, which assesses whether significant differences exist among the dataset sizes.

\begin{table*}[!tb]
\centering
\caption{Total energy (\si{\joule}) for NN use. Mean and standard deviation calculated by Eqs.~\eqref{eq:log_mean} and \eqref{eq:log_std}, the parameters $\mu$ and $\sigma$ in Eq.~\eqref{eq:lognormal}, and NMSE.}
\label{tb:NN_use_total}
\vspace{-0.5em}
\scriptsize{
\begin{tabular}{lrrrrrrrrrrrrrrr}
\toprule
&\multicolumn{5}{c}{M\'{a}laga}&\multicolumn{5}{c}{Stockholm}&\multicolumn{5}{c}{Paris}\\
\cmidrule(lr){2-6}\cmidrule(lr){7-11}\cmidrule(lr){12-16}
\# dataset&$E$&$SD$&$\mu$&$\sigma$&NMSE&$E$&$SD$&$\mu$&$\sigma$&NMSE&$E$&$SD$&$\mu$&$\sigma$&NMSE\\
\midrule
128 & 3.46 & 0.08 & 1.24 & 0.02 & $ 1.95 \times 10^{-1} $ & 3.43 & 0.08 & 1.23 & 0.02 & $ 2.09 \times 10^{-1} $ & 3.44 & 0.10 & 1.24 & 0.03 & $ 2.10 \times 10^{-1} $ \\
256 & 3.30 & 0.30 & 1.19 & 0.09 & $ 7.48 \times 10^{-1} $ & \textbf{2.71} & 0.07 & 1.00 & 0.03 & $ 1.37 \times 10^{-1} $ & 2.73 & 0.07 & 1.01 & 0.03 & $ 1.42 \times 10^{-1} $ \\
512 & 2.69 & 0.07 & 0.99 & 0.03 & $ 1.44 \times 10^{-1} $ & 2.72 & 0.07 & 1.00 & 0.03 & $ 1.32 \times 10^{-1} $ & 2.73 & 0.07 & 1.00 & 0.03 & $ 1.44 \times 10^{-1} $ \\
1024 & 2.68 & 0.07 & 0.99 & 0.03 & $ 1.46 \times 10^{-1} $ & 2.74 & 0.07 & 1.01 & 0.03 & $ 1.31 \times 10^{-1} $ & 2.73 & 0.07 & 1.00 & 0.03 & $ 1.36 \times 10^{-1} $ \\
2048 & 2.68 & 0.07 & 0.98 & 0.03 & $ 1.40 \times 10^{-1} $ & 2.72 & 0.07 & 1.00 & 0.03 & $ 1.02 \times 10^{-1} $ & 2.72 & 0.07 & 1.00 & 0.03 & $ 1.06 \times 10^{-1} $ \\
4096 & 2.66 & 0.07 & 0.98 & 0.03 & $ 1.11 \times 10^{-1} $ & 2.73 & 0.07 & 1.00 & 0.03 & $ 1.10 \times 10^{-1} $ & \textbf{2.72} & 0.07 & 1.00 & 0.03 & $ 1.06 \times 10^{-1} $ \\
8192 & \textbf{2.64} & 0.07 & 0.97 & 0.03 & $ 9.68 \times 10^{-2} $ & 2.72 & 0.07 & 1.00 & 0.03 & $ 9.62 \times 10^{-2} $ & 2.73 & 0.07 & 1.00 & 0.03 & $ 9.64 \times 10^{-2} $ \\
\midrule
$p$-value&\multicolumn{5}{l}{$<0.01$}&\multicolumn{5}{l}{$<0.01$}&\multicolumn{5}{l}{$<0.01$}\\
\bottomrule
\end{tabular}
}
\vspace{1em}
\centering
\caption{Execution time (\si{\milli\second}) for NN use. Mean and standard deviation calculated by Eqs.~\eqref{eq:log_mean} and \eqref{eq:log_std}, the parameters $\mu$ and $\sigma$ in Eq.~\eqref{eq:lognormal}, and NMSE.}
\label{tb:NN_use_time}
\vspace{-0.5em}
\scriptsize{
\begin{tabular}{lrrrrrrrrrrrrrrr}
\toprule
&\multicolumn{5}{c}{M\'{a}laga}&\multicolumn{5}{c}{Stockholm}&\multicolumn{5}{c}{Paris}\\
\cmidrule(lr){2-6}\cmidrule(lr){7-11}\cmidrule(lr){12-16}
\# dataset&$E$&$SD$&$\mu$&$\sigma$&NMSE&$E$&$SD$&$\mu$&$\sigma$&NMSE&$E$&$SD$&$\mu$&$\sigma$&NMSE\\
\midrule
128 & \textbf{63.81} & 1.47 & 4.16 & 0.02 & $ 3.34 \times 10^{-1} $ & \textbf{63.45} & 1.45 & 4.15 & 0.02 & $ 3.65 \times 10^{-1} $ & \textbf{64.08} & 1.52 & 4.16 & 0.02 & $ 3.43 \times 10^{-1} $ \\
256 & 64.10 & 2.61 & 4.16 & 0.04 & $ 5.44 \times 10^{-1} $ & 69.28 & 1.65 & 4.24 & 0.02 & $ 3.77 \times 10^{-1} $ & 69.70 & 1.67 & 4.24 & 0.02 & $ 3.90 \times 10^{-1} $ \\
512 & 68.92 & 1.66 & 4.23 & 0.02 & $ 3.63 \times 10^{-1} $ & 69.57 & 1.63 & 4.24 & 0.02 & $ 3.69 \times 10^{-1} $ & 69.47 & 1.65 & 4.24 & 0.02 & $ 3.65 \times 10^{-1} $ \\
1024 & 68.78 & 1.62 & 4.23 & 0.02 & $ 3.61 \times 10^{-1} $ & 69.58 & 1.66 & 4.24 & 0.02 & $ 3.63 \times 10^{-1} $ & 69.24 & 1.64 & 4.24 & 0.02 & $ 3.60 \times 10^{-1} $ \\
2048 & 68.64 & 1.62 & 4.23 & 0.02 & $ 3.56 \times 10^{-1} $ & 68.95 & 1.67 & 4.23 & 0.02 & $ 2.44 \times 10^{-1} $ & 68.79 & 1.65 & 4.23 & 0.02 & $ 2.51 \times 10^{-1} $ \\
4096 & 68.25 & 1.65 & 4.22 & 0.02 & $ 2.51 \times 10^{-1} $ & 69.19 & 1.68 & 4.24 & 0.02 & $ 2.63 \times 10^{-1} $ & 68.50 & 1.67 & 4.23 & 0.02 & $ 2.46 \times 10^{-1} $ \\
8192 & 67.85 & 1.64 & 4.22 & 0.02 & $ 1.99 \times 10^{-1} $ & 68.93 & 1.70 & 4.23 & 0.02 & $ 2.18 \times 10^{-1} $ & 68.80 & 1.67 & 4.23 & 0.02 & $ 2.17 \times 10^{-1} $ \\
\midrule
$p$-value&\multicolumn{5}{l}{$<0.01$}&\multicolumn{5}{l}{$<0.01$}&\multicolumn{5}{l}{$<0.01$}\\
\bottomrule
\end{tabular}
}
\vspace{1em}
\centering
\caption{Memory usage (\si{\kilo\byte}) for NN use. Mean and standard deviation calculated by Eqs.~\eqref{eq:log_mean} and \eqref{eq:log_std}, the parameters $\mu$ and $\sigma$ in Eq.~\eqref{eq:lognormal}, and NMSE.}
\label{tb:NN_use_memory}
\vspace{-0.5em}
\scriptsize{
\begin{tabular}{lrrrrrrrrrrrrrrr}
\toprule
&\multicolumn{5}{c}{M\'{a}laga}&\multicolumn{5}{c}{Stockholm}&\multicolumn{5}{c}{Paris}\\
\cmidrule(lr){2-6}\cmidrule(lr){7-11}\cmidrule(lr){12-16}
\# dataset&$E$&$SD$&$\mu$&$\sigma$&NMSE&$E$&$SD$&$\mu$&$\sigma$&NMSE&$E$&$SD$&$\mu$&$\sigma$&NMSE\\
\midrule
128 & 117.68 & 4.29 & 4.77 & 0.04 & $ 3.17 \times 10^{-1} $ & 121.95 & 4.35 & 4.80 & 0.04 & $ 3.07 \times 10^{-1} $ & \textbf{122.07} & 3.85 & 4.80 & 0.03 & $ 2.69 \times 10^{-1} $ \\
256 & 117.66 & 4.31 & 4.77 & 0.04 & $ 3.22 \times 10^{-1} $ & \textbf{121.85} & 3.85 & 4.80 & 0.03 & $ 2.71 \times 10^{-1} $ & 122.08 & 3.85 & 4.80 & 0.03 & $ 2.73 \times 10^{-1} $ \\
512 & 117.67 & 4.22 & 4.77 & 0.04 & $ 3.16 \times 10^{-1} $ & 121.95 & 4.22 & 4.80 & 0.03 & $ 3.13 \times 10^{-1} $ & 122.20 & 4.35 & 4.80 & 0.04 & $ 3.22 \times 10^{-1} $ \\
1024 & 117.72 & 4.11 & 4.77 & 0.03 & $ 2.80 \times 10^{-1} $ & 121.93 & 4.18 & 4.80 & 0.03 & $ 3.04 \times 10^{-1} $ & 122.14 & 4.26 & 4.80 & 0.03 & $ 3.14 \times 10^{-1} $ \\
2048 & \textbf{117.65} & 3.80 & 4.77 & 0.03 & $ 2.61 \times 10^{-1} $ & 121.99 & 4.27 & 4.80 & 0.04 & $ 3.13 \times 10^{-1} $ & 122.13 & 4.23 & 4.80 & 0.03 & $ 3.10 \times 10^{-1} $ \\
4096 & 117.67 & 3.80 & 4.77 & 0.03 & $ 2.63 \times 10^{-1} $ & 121.98 & 4.21 & 4.80 & 0.03 & $ 3.05 \times 10^{-1} $ & 122.18 & 4.17 & 4.80 & 0.03 & $ 3.02 \times 10^{-1} $ \\
8192 & 117.71 & 4.23 & 4.77 & 0.04 & $ 3.10 \times 10^{-1} $ & 121.99 & 4.12 & 4.80 & 0.03 & $ 2.96 \times 10^{-1} $ & 122.14 & 4.22 & 4.80 & 0.03 & $ 3.07 \times 10^{-1} $ \\
\midrule
$p$-value&\multicolumn{5}{l}{$<0.01$}&\multicolumn{5}{l}{$<0.01$}&\multicolumn{5}{l}{$<0.01$}\\
\bottomrule
\end{tabular}
}
\vspace{-2em}
\end{table*}
The results in Tables~\ref{tb:NN_use_total}--\ref{tb:NN_use_memory} confirm that energy consumption decreases as the training dataset size increases across all instances. The Kruskal-Wallis test supports this observation, showing statistically significant differences at the 1\% level. This suggests that larger training datasets may contribute to more efficient NN prediction, reducing the computational cost of NN use. In surrogate-assisted search, where NN is used multiple times, even slight reductions in energy consumption can have a substantial cumulative impact.

While execution time shows significant differences across dataset sizes, no clear decreasing trend is observed. Similarly, memory usage remains comparable across dataset sizes, suggesting that NN utilization exhibits near-constant execution time and memory demands regardless of the training dataset size. One possible explanation is that once an NN is trained, its structure remains fixed, and inference relies on a predetermined set of parameters and operations, making its computational cost largely independent of the dataset used for training.

Comparing instances, energy consumption is lowest for \Malaga, followed by Stockholm, and highest for Paris, which can be attributed to the increased computational load due to more NN parameters. This trend is consistent with the observations in NN training, where both higher input dimensionality and a larger number of network parameters contribute to greater energy requirements. Similarly, execution time is slightly lower for \Malaga than Stockholm and Paris, suggesting that the computational cost of NN use increases with network complexity, albeit to a lesser extent than in training. Regarding memory consumption, while no notable differences are observed between Stockholm and Paris, the \Malaga instance exhibits lower memory usage due to its reduced number of NN parameters.

We compare the computational cost of NN use with that of evaluating tentative solutions. In terms of energy consumption, NN use consistently results in lower energy usage than solution evaluations across all instances, achieving reductions of over 98\% even in the most demanding cases. A similar trend is observed for execution time, where NN use is faster in all instances, reducing computation time by more than 98\% in the worst case. The reduction in memory usage is even more substantial, with NN use achieving over 99.7\% savings compared to solution evaluations. These results demonstrate the significant advantage of surrogate models in reducing the computational cost of solution evaluations.

We further conducted pairwise statistical comparisons across different dataset sizes to clarify the effect of dataset size on computational performance. While the previous analysis confirmed a general trend of decreasing energy, it is essential to assess the statistical significance of these differences to determine whether they consistently hold across instances. 

\begin{figure}[!tb]
\begin{minipage}[t]{0.32\columnwidth}
\includegraphics[width=\textwidth]{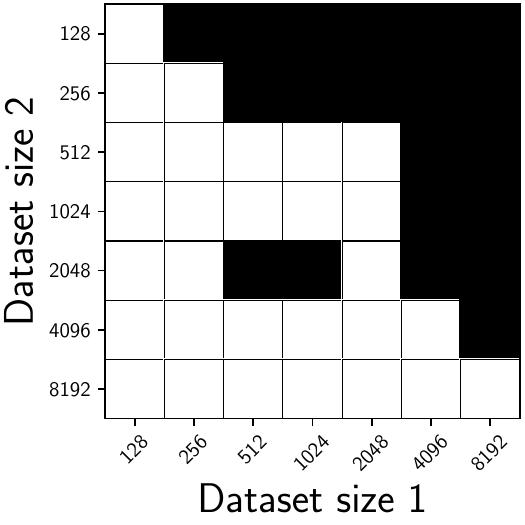}
\subcaption{\Malaga}
\end{minipage}
\begin{minipage}[t]{0.32\columnwidth}
\includegraphics[width=\textwidth]{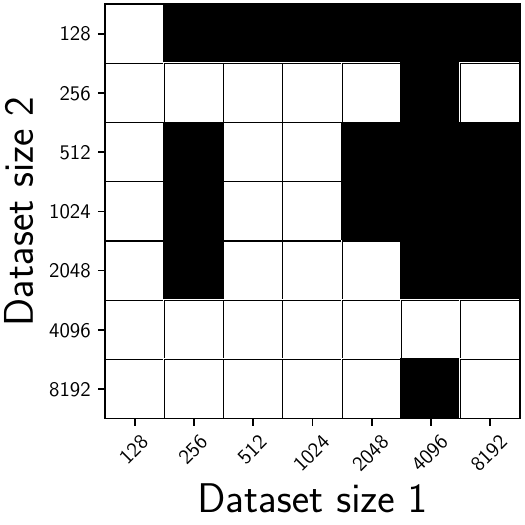}
\subcaption{Stockholm}
\end{minipage}
\begin{minipage}[t]{0.32\columnwidth}
\includegraphics[width=\textwidth]{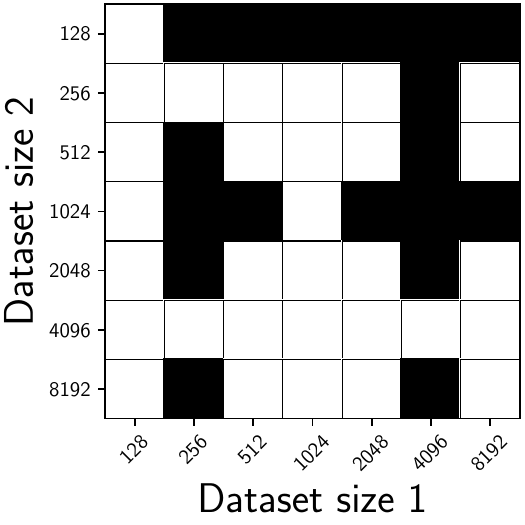}
\subcaption{Paris}
\end{minipage}
\vspace{-0.5em}
\caption{Energy consumption: Post-hoc analysis results of pairwise statistical comparisons across different dataset sizes}
\label{fig:sig_diff_NN_use_Total}
\vspace{-1.5em}
\end{figure}
Fig.~\ref{fig:sig_diff_NN_use_Total} presents the results of post-hoc pairwise statistical tests for energy consumption across different dataset sizes. Each cell represents a one-sided comparison between two dataset sizes. A black cell indicates that the energy consumption for ``Dataset size 1'' is significantly lower than that for ``Dataset size 2'' at a 1\% significance level. In contrast, a white cell indicates that "Dataset size 1" is not significantly lower or significantly higher than ``Dataset size 2.'' The lower triangular part of the matrix reflects the general trend of increasing energy consumption with larger dataset sizes, while the upper triangular part corresponds to comparisons where larger datasets result in reduced energy.

Fig.~\ref{fig:sig_diff_NN_use_Total} indicates that significant differences are consistently observed in the upper triangular part for all instances. Then, energy consumption decreases significantly as the dataset size increases, suggesting that larger training datasets contribute to reduced computational costs during NN utilization.

\begin{table}[!tb]
\centering
\caption{MAPE [\%] of trained NNs}
\label{tb:mape_NN}
\vspace{-0.5em}
\scriptsize{
\begin{tabular}{lrrr}
\toprule
\# dataset & \Malaga & Stockholm & Paris \\
\midrule
128 & 37.31 & 41.01 & 21.73 \\
256 & 33.60 & 36.82 & 19.63 \\
512 & 30.92 & 38.60 & 26.57 \\
1024 & 31.87 & 33.28 & 19.21 \\
2048 & 29.85 & 33.62 & 18.71 \\
4096 & 27.98 & 34.30 & 18.10 \\
8192 & \textbf{25.27} & \textbf{32.00} & \textbf{16.10} \\
\midrule
$p$-value&$<0.01$&$<0.01$&$<0.01$\\
\bottomrule
\end{tabular}
}
\vspace{-2em}
\end{table}

\begin{figure*}[!tb]
\centering
\centering
\includegraphics[width=0.7\linewidth]{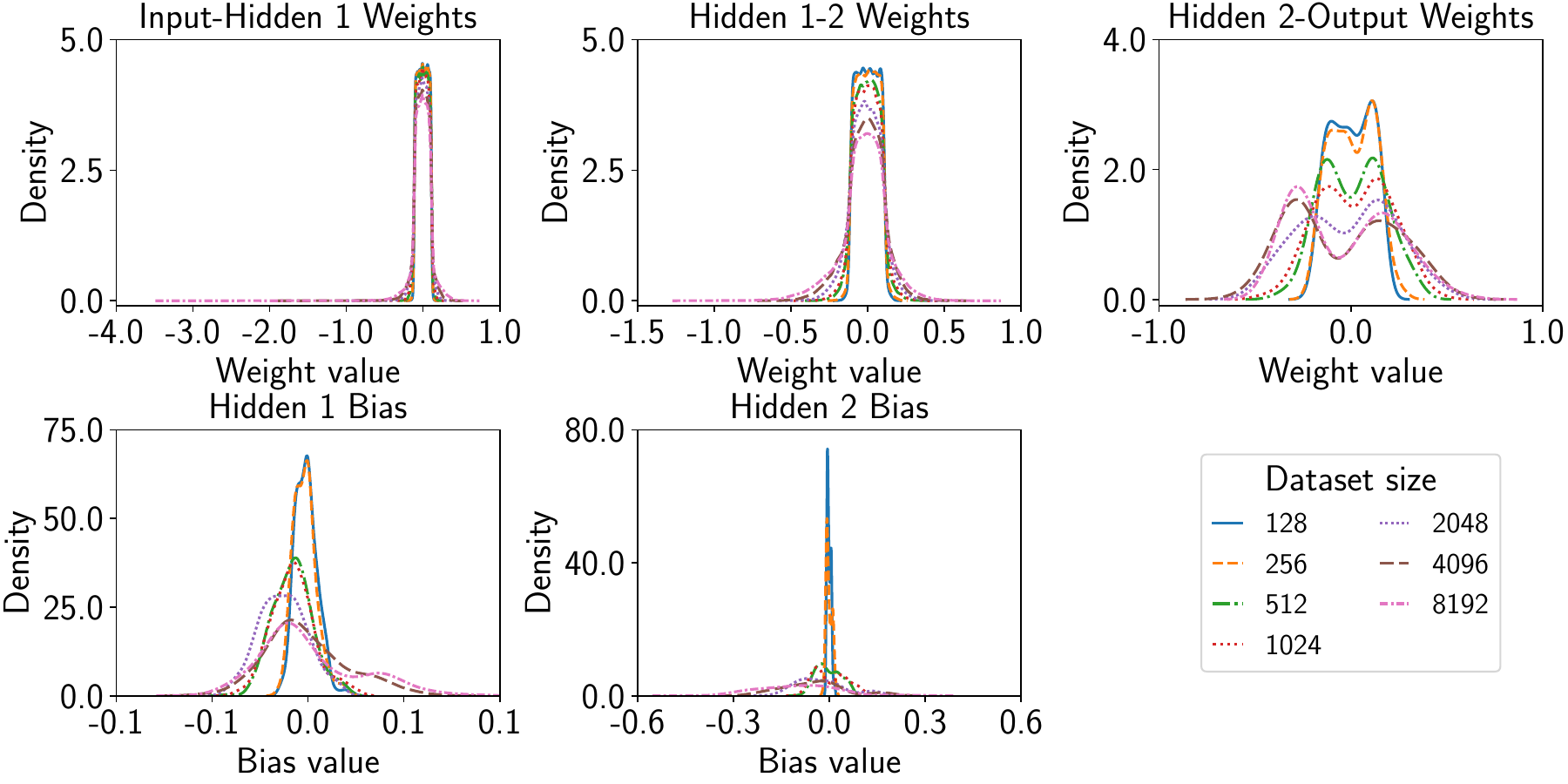}
\vspace{-0.5em}
\caption{Distribution of weights and biases for each layer across various dataset sizes on the \Malaga instance}
\label{fig:NN_weights}
\vspace{-1.5em}
\end{figure*}
To assess the prediction accuracy of trained NNs, the mean absolute percentage error (MAPE) is computed based on the actual and predicted objective function value pairs as:
\begin{equation}
    \mathrm{MAPE} = \frac{1}{n} \sum_{i=1}^n \left| \frac{\hat{y}_i - y_i}{y_i} \right| \times 100,
\end{equation}
where $y_i$ and $\hat{y}_i$ represent the actual and predicted objective values for $n$ testing data points. A smaller MAPE value indicates a smaller difference between actual and predicted fitness values, with MAPE approaching 0 indicating a closer match.
Table~\ref{tb:mape_NN} presents the MAPE values for each instance and training dataset size. The bottom row reports the $p$-values from the Kruskal-Wallis test, which assesses the significance of differences in prediction accuracy across different dataset sizes.

As shown in Table~\ref{tb:mape_NN}, prediction accuracy improves as the dataset size increases. Specifically, comparing sizes 128 and 8192, the MAPE improves by approximately 9\% for \Malaga and Stockholm, and by about 5\% for Paris.
In all instances, the highest prediction accuracy (lowest MAPE) is consistently achieved when using the largest training dataset size (8192), highlighting the benefit of larger training sets. 

These results indicate that NN surrogates achieve both higher accuracy and lower energy consumption with larger training datasets. To investigate the cause of this energy reduction, the next subsection analyzes the internal structure of the trained networks and how it relates to computational efficiency.

\subsection{Delving into the Internal Structure of the Trained NN}\label{sec:NN_structure}
Fig.~\ref{fig:NN_weights} presents the distribution of weights and biases for each layer across various dataset sizes. The horizontal axis represents the values of weights or biases, and the vertical axis indicates the density of the data. Each line shows a kernel density estimate (KDE) plot for a specific dataset size. 

The results demonstrate that the weight distributions become wider as the training dataset size increases. For example, regarding the weights from the second hidden layer to the output (upper right plot), the weights are concentrated near zero when the dataset size is small, but spread out over a wider range for larger dataset sizes, indicating greater variability in the learned parameters. Similarly, the biases (lower plots) shift toward negative values with larger dataset sizes, deviating from their initial concentration around zero when trained on smaller datasets. These analyses show that weight distributions expand and biases shift as the training dataset size becomes rich. 

\begin{table*}[!tb]
\centering
\caption{The percentage of zero outputs for the trained NNs}
\label{tb:zero_ratio}
\vspace{-0.5em}
\scriptsize{
\begin{tabular}{lrrrrrrrrrrrr}
\toprule
&\multicolumn{6}{c}{First hidden layer}&\multicolumn{6}{c}{Second hidden layer}\\
\cmidrule(lr){2-7}\cmidrule(lr){8-13}
&\multicolumn{2}{c}{\Malaga}&\multicolumn{2}{c}{Stockholm}&\multicolumn{2}{c}{Paris}&\multicolumn{2}{c}{\Malaga}&\multicolumn{2}{c}{Stockholm}&\multicolumn{2}{c}{Paris}\\
\# dataset&Avg.&Stdv.&Avg.&Stdv.&Avg.&Stdv.&Avg.&Stdv.&Avg.&Stdv.&Avg.&Stdv.\\
\midrule
128&57.89&2.68&67.86&2.09&70.43&1.86&58.45&3.73&63.85&2.54&69.03&2.60\\
256&65.62&2.75&79.64&2.28&78.41&1.73&63.68&3.02&69.01&2.33&75.58&2.51\\
512&88.76&2.13&95.89&2.27&93.11&2.87&68.70&4.42&74.42&4.87&79.48&2.87\\
1024&89.92&2.03&98.34&0.58&97.71&1.42&68.77&2.70&70.03&3.97&79.30&3.50\\
2048&\textbf{92.96}&1.33&98.85&0.39&98.92&0.35&75.14&2.19&70.99&3.48&77.96&3.17\\
4096&91.98&1.25&\textbf{98.79}&0.31&\textbf{99.11}&0.28&83.52&2.04&78.76&2.98&83.68&1.91\\
8192&91.54&1.24&98.21&0.40&98.68&0.33&\textbf{90.18}&1.37&\textbf{87.09}&2.26&\textbf{90.11}&1.55\\
\bottomrule
\end{tabular}
}
\vspace{-2em}
\end{table*}
\begin{figure}[tb]
\begin{minipage}{0.49\columnwidth}
\centering
\includegraphics[width=\columnwidth]{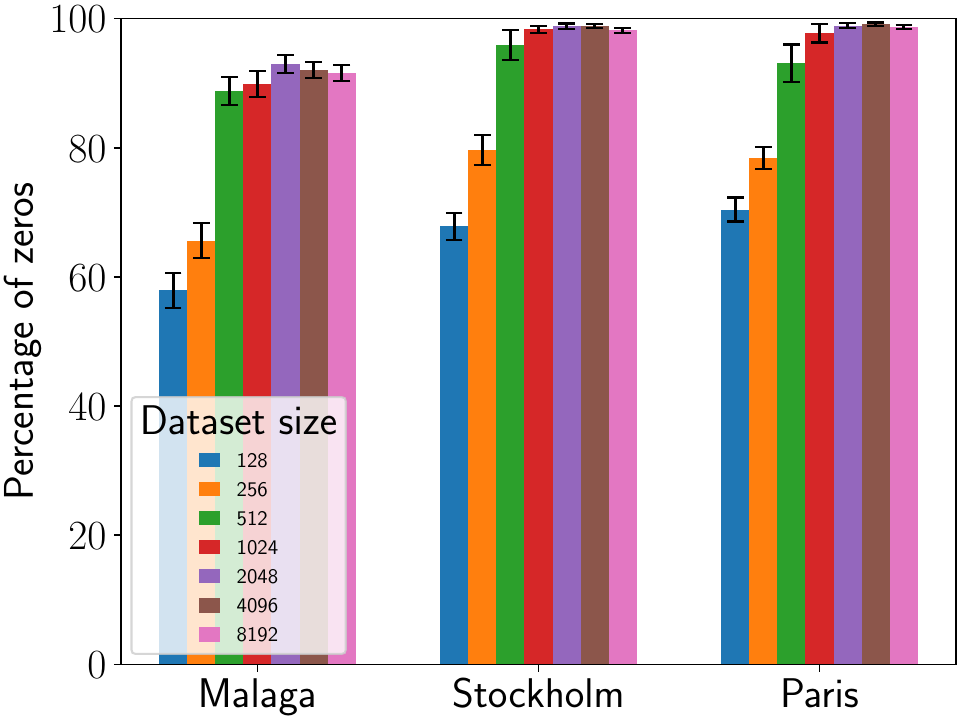}
\vspace{-1em}
\caption{The percentage of the first hidden layer outputs that are zero}
\label{fig:zero_ratio1}
\end{minipage}
\hfill
\begin{minipage}{0.49\columnwidth}
\centering
\includegraphics[width=\columnwidth]{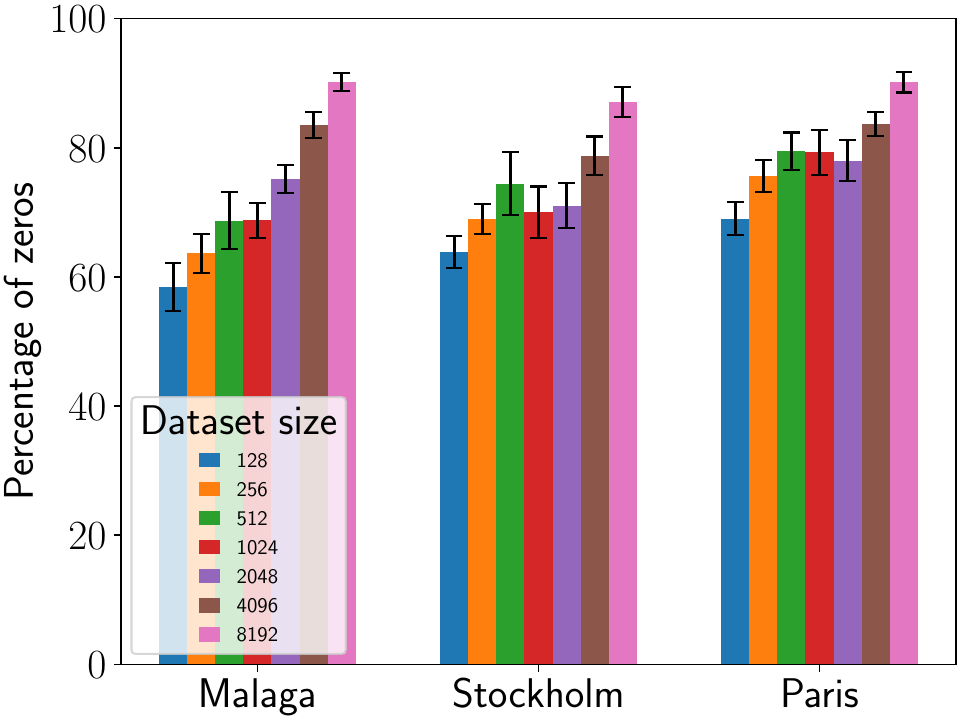}
\vspace{-1em}
\caption{The percentage of the second hidden layer outputs that are zero}
\label{fig:zero_ratio2}
\end{minipage}
\vspace{-1.5em}
\end{figure}
To further examine the impact of these changes on NN behavior, Table~\ref{tb:zero_ratio} summarizes the mean and standard deviation of the percentage of zero outputs, highlighting the largest value in bold. In addition, Figs.~\ref{fig:zero_ratio1} and \ref{fig:zero_ratio2} present the percentage of zero outputs in the first and second hidden layers, respectively. The horizontal axis represents each instance, and the vertical axis shows the proportion of zero outputs.  Since ReLU activation is used, negative inputs result in zero outputs. 

These analyses reveal that NNs trained with larger datasets exhibit a higher ratio of zero outputs in their hidden layers. Specifically, for the first hidden layer, the ratio of zero outputs ranges from 57\% (\Malaga) to 70\% (Paris) when the dataset size is 128, and increases to 65\%--79\% for a dataset size of 256. In contrast, when the dataset size exceeds 512, the percentage exceeds 88\% in \Malaga and 93\% in both Stockholm and Paris. In the second hidden layer, the ratio of zero outputs also increases with the dataset size, although the increase is more gradual. This increase in zero outputs reduces the number of non-zero multiplications in subsequent layers, thereby contributing to lower energy consumption. These findings suggest that training NNs with sufficiently large datasets results in broader weight distributions and increased sparsity in hidden layer activations, both of which enhance computational efficiency. The behavioral shift observed around a dataset size of 512 is consistent with the trends shown in Table~\ref{tb:NN_use_total}, indicating that well-trained NNs using large datasets can significantly reduce energy consumption during utilization.

Based on the above discussion, it has been demonstrated that NNs trained with large datasets offer advantages not only in terms of prediction accuracy but also in terms of energy consumption. The next subsection discusses how these advantages translate into benefits in surrogate-assisted optimization.

\subsection{Long-term Energy Benefits of Large-Dataset Training}\label{sec:NN_energy_benefits}
Training NNs with larger datasets requires higher initial energy consumption but offers substantial long-term benefits in surrogate-assisted optimization by reducing the energy per surrogate use. For example, in the \Malaga instance, training with a dataset of 128 consumes $E_\mathrm{train}^{128}=\SI{217.79}{\joule}$, and each surrogate use requires $E_\mathrm{use}^{128}=\SI{3.46}{\joule}$. In contrast, training with a dataset of 8192 requires $E_\mathrm{train}^{8192}=\SI{2574.66}{\joule}$, while reducing the energy per surrogate use to $E_\mathrm{use}^{8192}=\SI{2.64}{\joule}$. Thus, each surrogate use with the larger dataset saves \SI{0.82}{\joule} compared to the small dataset.

Although the initial training energy for the larger dataset is substantially higher (an additional \SI{2356.87}{\joule}), this investment is gradually offset as the surrogate is repeatedly used. Specifically, if the NN is trained once and then used $N$ times, the break-even point---where the total energy consumption of the large dataset becomes lower than that of the small dataset---occurs after approximately:
\begin{equation}
N = \frac{E_\mathrm{train}^{8192} - E_\mathrm{train}^{128}}{E_\mathrm{use}^{128} - E_\mathrm{use}^{8192}} = \frac{2356.87}{0.82} \approx 2874
\end{equation}
uses. Beyond this threshold, the overall energy cost of using a well-trained NN with a large dataset becomes lower than that of using a less accurate NN trained with a small dataset. This trade-off is relevant in surrogate-assisted optimization, where thousands or tens of thousands of NN uses are typically performed. In such scenarios, the cumulative energy savings from using a large training dataset can significantly outweigh the initial training cost, making large dataset training a practical and energy-efficient strategy for long-running optimization tasks.

These results indicate that a well-trained surrogate offers dual benefits in surrogate-assisted optimization. Beyond providing more accurate predictions, it contributes to reducing overall energy consumption during optimization. This suggests that investing in larger training datasets not only improves optimization performance but also enhances energy efficiency, highlighting the critical role of surrogate design in achieving both high performance and sustainable, resource-efficient optimization.

\section{Algorithm Execution}\label{sec:algorithm_execution}
This section presents three experiments designed to evaluate the energy consumption, execution time, and search performance of SAPSO and SAGA. These experiments address the research questions (RQs) as follows:
\begin{description}
    \setlength\parskip{-1em}%
    \setlength\itemsep{0em}%
\item[Experiment 1:] Evaluation of computational performance---energy consumption, execution time, and memory usage---to address RQ1 and RQ2.
    \setlength\parskip{0em}%
\item[Experiment 2:] Evaluation of optimization performance---the quality of solutions obtained---to address RQ3 and RQ4.
\item[Experiment 3:] Evaluation of surrogate model accuracy by comparing predicted and actual objective values---also related to RQ3 and RQ4.
\end{description}
\vspace{-1em}

\subsection{Parameter Settings}
\begin{table}[tb]
\centering
\caption{Parameter settings for PSO and GA families}
\label{tb:params}
\vspace{-0.5em}
\scriptsize{
\begin{tabular}{llr}
\toprule
&Parameter&Value\\
\midrule
\multirow{5}{*}{PSO}&Local coefficient ($\phi_1$)&2.05\\
&Global coefficient ($\phi_2$)&2.05\\
&Velocity truncation factor ($\lambda$)&0.5\\
&Maximum inertia weight ($w_{max}$)&0.5\\
&Minimum inertia weight ($w_{min}$)&0.1\\
\midrule
\multirow{3}{*}{GA}&Crossover rate ($p_c$)&1.0\\
&Mutation rate ($p_m$)&$1/D$\\
&Mut. distribution index&20.0\\
\midrule
\multirow{4}{*}{Common}&Population (Swarm) size ($N$)&100\\
&Maximum evaluations ($MaxFE$)&30000\\
&The number of initial samples ($N_t$)&100, 8192\\
&The number of re-evaluations ($N_r$)&10\\
\bottomrule
\end{tabular}
}
\vspace{-2em}
\end{table}
Table~\ref{tb:params} summarizes the parameter values used for PSO and GA families. The parameter settings follow the configurations used in previous studies~\citep{GARCIANIETO2012274,Segredo2019}.

For the PSO settings, both the local and global acceleration coefficients ($\phi_1$ and $\phi_2$ in Eq.~\eqref{eq:update_velocity}) are set to 2.05. The velocity truncation factor $\lambda$ is set to 0.5, meaning that the ceiling and floor functions in Eq.~\eqref{eq:velocity_calculation} are selected equally. The inertia weight decreases linearly from 0.5 to 0.1 according to Eq.~\eqref{eq:weight_schedule}.

For the GA settings, selection is performed using binary tournament selection, and uniform crossover is applied with a crossover rate of 1.0. Mutation is implemented using integer polynomial mutation~\citep{deb1996combined}, with the mutation rate set to $1/D$, where $D$ is the number of design variables. The mutation distribution index is set to 20.0. These settings were chosen to ensure comparability with the PSO-based configurations.

All algorithms share common parameters for the population (or swarm) size $N=100$ and the maximum number of evaluations $MaxFE=30000$.
For both SAPSO and SAGA, the size of the initial training dataset $N_t$ is set to either 100 or 8200. The smaller size ($N_t=100$) represents a minimal setting where the surrogate is trained using the initial randomly generated population. The larger size ($N_t=8200$) approximates the full dataset of 8192 samples used in preliminary experiments, which provided sufficient accuracy for the NN.

To distinguish surrogate-assisted configurations, we denote the algorithms using the small dataset as SAPSO-ps, SAPSO-rs, SAGA-ps, and SAGA-rs, respectively, and those using the large dataset as SAPSO-pl, SAPSO-rl, SAGA-pl, and SAGA-rl. The suffixes ``ps'' and ``pl'' refer to pre-training with small and large datasets, while ``rs'' and ``rl'' refer to retraining using small and large datasets. For both SAPSO-r and SAGA-r configurations, the number of re-evaluations $N_r$ is set to 10. Each algorithm was executed independently 11 times to ensure statistical reliability.

\subsection{Experiment 1: Energy Consumption, Time, and Memory}

\begin{table*}[!tb]
\caption{\Malaga instance: Average energy consumption, execution time, and their standard deviation for each optimization component}
\label{tab:measurement_summary_malaga}
\vspace{-0.5em}
\centering
\scriptsize{
\begin{tabular}{llrrrrrrrrrrrr}
\toprule
 &  & \multicolumn{2}{c}{Evaluation} & \multicolumn{2}{c}{Training} & \multicolumn{2}{c}{Use} & \multicolumn{2}{c}{Initialization} & \multicolumn{2}{c}{Update} & \multicolumn{2}{c}{Total} \\
 \cmidrule(lr){3-4}\cmidrule(lr){5-6}\cmidrule(lr){7-8}\cmidrule(lr){9-10}\cmidrule(lr){11-12}\cmidrule(lr){13-14}
 & Measure & Avg. & Stdv. & Avg. & Stdv. & Avg. & Stdv. & Avg. & Stdv. & Avg. & Stdv. & Avg. & Stdv. \\
\midrule
\multirow[c]{3}{*}{PSO} & CPU (\si{\joule}) & 4268196.53 & 131835.40 & NA & NA & NA & NA & 0.97 & 0.05 & 419.39 & 13.95 & 4268616.90 & 131845.85 \\
 & DRAM (\si{\joule}) & 234512.72 & 18539.39 & NA & NA & NA & NA & \textbf{0.05} & 0.01 & 56.14 & 6.20 & 234568.90 & 18545.48 \\
 & Time (\si{\second}) & 83951.11 & 1638.43 & NA & NA & NA & NA & 0.02 & 0.00 & 13.76 & 0.33 & 83964.89 & 1638.56 \\
\midrule
\multirow[c]{3}{*}{SAPSO-ps} & CPU (\si{\joule}) & \textbf{21343.90} & 559.77 & \textbf{317.95} & 2.09 & 99855.55 & 5604.39 & 0.96 & 0.04 & 442.90 & 4.60 & \textbf{121961.26} & 5088.79 \\
 & DRAM (\si{\joule}) & \textbf{1187.37} & 136.15 & \textbf{16.73} & 3.21 & 5301.92 & 1855.97 & 0.05 & 0.01 & 22.37 & 8.03 & \textbf{6528.44} & 1911.39 \\
 & Time (\si{\second}) & \textbf{417.96} & 5.11 & \textbf{6.47} & 0.21 & 2156.34 & 121.89 & \textbf{0.02} & 0.00 & \textbf{6.66} & 0.23 & \textbf{2587.44} & 125.54 \\
\midrule
\multirow[c]{3}{*}{SAPSO-pl} & CPU (\si{\joule}) & 1205672.26 & 36607.33 & 2518.65 & 179.32 & 63919.30 & 2339.78 & \textbf{0.95} & 0.05 & \textbf{412.44} & 18.22 & 1272523.61 & 38870.20 \\
 & DRAM (\si{\joule}) & 67643.71 & 5852.99 & 106.85 & 20.71 & 1834.77 & 380.36 & 0.05 & 0.01 & 22.78 & 1.86 & 69608.16 & 5716.56 \\
 & Time (\si{\second}) & 23683.85 & 430.02 & 89.67 & 8.15 & 1657.52 & 15.89 & 0.02 & 0.00 & 8.55 & 0.18 & 25439.62 & 444.01 \\
\midrule
\multirow[c]{3}{*}{SAPSO-rs} & CPU (\si{\joule}) & 460367.91 & 13984.06 & 214141.85 & 8588.01 & 90028.60 & 1374.65 & 0.97 & 0.06 & 428.13 & 14.96 & 764967.45 & 21823.16 \\
 & DRAM (\si{\joule}) & 26051.96 & 1300.95 & 8621.75 & 616.80 & 3293.86 & 244.42 & 0.05 & 0.01 & \textbf{14.59} & 1.25 & 37982.21 & 1920.38 \\
 & Time (\si{\second}) & 9094.34 & 218.81 & 4915.96 & 233.13 & 2237.05 & 38.93 & 0.02 & 0.00 & 6.90 & 0.09 & 16254.26 & 385.46 \\
\midrule
\multirow[c]{3}{*}{SAPSO-rl} & CPU (\si{\joule}) & 1508854.46 & 46230.83 & 607843.73 & 37647.45 & \textbf{62951.00} & 2079.81 & 0.95 & 0.05 & 418.92 & 12.95 & 2180069.06 & 82562.70 \\
 & DRAM (\si{\joule}) & 82620.57 & 4424.22 & 32251.17 & 6123.06 & \textbf{1832.22} & 344.00 & 0.05 & 0.01 & 23.35 & 1.30 & 116727.36 & 4319.05 \\
 & Time (\si{\second}) & 29761.34 & 550.28 & 24470.52 & 1700.18 & \textbf{1632.97} & 14.73 & 0.02 & 0.00 & 8.77 & 0.15 & 55873.62 & 2070.84 \\
 \midrule\midrule
 \multirow[c]{3}{*}{GA} & CPU (\si{\joule}) & 3831097.79 & 70541.54 & NA & NA & NA & NA & \textbf{0.68} & 0.05 & 707.18 & 9.42 & 3831805.66 & 70541.78 \\
 & DRAM (\si{\joule}) & 216640.88 & 24865.41 & NA & NA & NA & NA & \textbf{0.03} & 0.02 & 100.37 & 16.59 & 216741.28 & 24881.65 \\
 & Time (\si{\second}) & 75362.86 & 767.22 & NA & NA & NA & NA & 0.02 & 0.01 & 29.75 & 1.44 & 75392.63 & 767.06 \\
 \midrule
 \multirow[c]{3}{*}{SAGA-ps} & CPU (\si{\joule}) & \textbf{21302.30} & 498.29 & \textbf{317.09} & 3.28 & 99306.40 & 5458.06 & 0.72 & 0.04 & 686.33 & 13.42 & \textbf{121612.84} & 4998.54 \\
 & DRAM (\si{\joule}) & \textbf{1133.83} & 115.43 & \textbf{15.75} & 3.74 & 5372.08 & 1683.31 & 0.03 & 0.01 & 36.38 & 11.46 & \textbf{6558.07} & 1795.39 \\
 & Time (\si{\second}) & \textbf{416.55} & 4.84 & \textbf{6.57} & 0.45 & 2144.16 & 123.64 & 0.01 & 0.00 & \textbf{12.80} & 0.49 & \textbf{2580.11} & 127.50 \\
 \midrule
\multirow[c]{3}{*}{SAGA-pl} & CPU (\si{\joule}) & 1124170.74 & 25274.93 & 2303.16 & 748.53 & 63360.61 & 2390.43 & 0.71 & 0.03 & \textbf{662.09} & 23.32 & 1190497.32 & 27661.10 \\
 & DRAM (\si{\joule}) & 64383.38 & 6979.02 & 105.85 & 22.51 & 1818.60 & 384.89 & 0.04 & 0.01 & 39.46 & 2.58 & 66347.32 & 6751.01 \\
 & Time (\si{\second}) & 22102.28 & 244.41 & 89.74 & 8.76 & 1643.44 & 18.75 & 0.01 & 0.00 & 17.52 & 0.39 & 23852.99 & 254.36 \\
 \midrule
\multirow[c]{3}{*}{SAGA-rs} & CPU (\si{\joule}) & 510434.09 & 25614.72 & 213669.47 & 8711.64 & 89480.30 & 1453.70 & 0.72 & 0.04 & 674.57 & 22.05 & 814259.15 & 33827.20 \\
 & DRAM (\si{\joule}) & 28642.68 & 2208.55 & 8290.05 & 753.21 & 3136.55 & 138.40 & 0.04 & 0.01 & \textbf{23.28} & 1.05 & 40092.60 & 2992.54 \\
 & Time (\si{\second}) & 10056.01 & 414.17 & 4905.68 & 230.48 & 2225.13 & 40.99 & \textbf{0.01} & 0.00 & 13.70 & 0.33 & 17200.54 & 609.08 \\
 \midrule
\multirow[c]{3}{*}{SAGA-rl} & CPU (\si{\joule}) & 1391533.91 & 29498.57 & 607235.54 & 36799.05 & \textbf{62582.96} & 2083.73 & 0.72 & 0.05 & 663.91 & 29.07 & 2062017.04 & 66713.14 \\
 & DRAM (\si{\joule}) & 77934.55 & 5061.09 & 29315.46 & 7377.73 & \textbf{1702.05} & 377.70 & 0.04 & 0.01 & 39.51 & 1.48 & 108991.61 & 5837.02 \\
 & Time (\si{\second}) & 27444.45 & 283.92 & 24426.45 & 1712.59 & \textbf{1624.75} & 15.17 & 0.01 & 0.00 & 17.82 & 0.40 & 53513.48 & 1869.29 \\
\bottomrule
\end{tabular}
}
\end{table*}
\begin{table*}[!tb]
\tabcolsep = 5.3pt
\caption{Stockholm instance: Average energy consumption, execution time, and their standard deviation for each optimization component}
\label{tab:measurement_summary_stockholm}
\vspace{-0.5em}
\centering
\scriptsize{
\begin{tabular}{llrrrrrrrrrrrr}
\toprule
 &  & \multicolumn{2}{c}{Evaluation} & \multicolumn{2}{c}{Training} & \multicolumn{2}{c}{Use} & \multicolumn{2}{c}{Initialization} & \multicolumn{2}{c}{Update} & \multicolumn{2}{c}{Total} \\
 \cmidrule(lr){3-4}\cmidrule(lr){5-6}\cmidrule(lr){7-8}\cmidrule(lr){9-10}\cmidrule(lr){11-12}\cmidrule(lr){13-14}
 & Measure & Avg. & Stdv. & Avg. & Stdv. & Avg. & Stdv. & Avg. & Stdv. & Avg. & Stdv. & Avg. & Stdv. \\
\midrule
\multirow[c]{3}{*}{PSO} & CPU (\si{\joule}) & 10660720.68 & 460615.10 & NA & NA & NA & NA & 1.71 & 0.10 & \textbf{428.60} & 20.41 & 10661150.99 & 460623.69 \\
 & DRAM (\si{\joule}) & 571168.05 & 55731.58 & NA & NA & NA & NA & \textbf{0.09} & 0.02 & 56.37 & 5.49 & 571224.51 & 55736.55 \\
 & Time (\si{\second}) & 209209.88 & 8112.83 & NA & NA & NA & NA & \textbf{0.03} & 0.00 & 14.01 & 0.57 & 209223.93 & 8112.84 \\
\midrule
\multirow[c]{3}{*}{SAPSO-ps} & CPU (\si{\joule}) & \textbf{54298.46} & 867.73 & \textbf{351.43} & 1.84 & 100194.09 & 5489.36 & 1.71 & 0.08 & 461.47 & 10.29 & \textbf{155307.16} & 4981.63 \\
 & DRAM (\si{\joule}) & \textbf{2991.04} & 239.50 & \textbf{20.73} & 3.19 & 5524.25 & 1623.09 & 0.10 & 0.01 & 25.14 & 6.80 & \textbf{8561.27} & 1815.46 \\
 & Time (\si{\second}) & \textbf{1066.04} & 12.11 & \textbf{6.96} & 0.21 & 2164.80 & 123.49 & 0.03 & 0.00 & \textbf{7.06} & 0.22 & \textbf{3244.89} & 120.43 \\
\midrule
\multirow[c]{3}{*}{SAPSO-pl} & CPU (\si{\joule}) & 3051057.54 & 97315.35 & 3811.30 & 1810.20 & 64159.60 & 2336.90 & \textbf{1.68} & 0.10 & 433.11 & 21.21 & 3119463.23 & 98170.17 \\
 & DRAM (\si{\joule}) & 163188.06 & 18399.99 & 327.25 & 13.68 & \textbf{2125.66} & 399.08 & 0.09 & 0.03 & 25.37 & 1.19 & 165666.42 & 18100.51 \\
 & Time (\si{\second}) & 59847.07 & 1767.74 & 83.47 & 2.22 & 1664.05 & 16.38 & 0.03 & 0.00 & 8.95 & 0.21 & 61603.58 & 1766.41 \\
\midrule
\multirow[c]{3}{*}{SAPSO-rs} & CPU (\si{\joule}) & 1129918.63 & 42126.15 & 350511.19 & 17133.91 & 89673.38 & 2275.64 & 1.72 & 0.07 & 444.59 & 20.53 & 1570549.51 & 48812.81 \\
 & DRAM (\si{\joule}) & 65076.73 & 2799.69 & 25111.19 & 706.41 & 3339.09 & 181.70 & 0.10 & 0.01 & \textbf{16.63} & 1.18 & 93543.74 & 3050.35 \\
 & Time (\si{\second}) & 22197.21 & 858.25 & 6613.71 & 103.01 & 2258.28 & 30.96 & 0.03 & 0.00 & 7.28 & 0.09 & 31076.51 & 845.08 \\
\midrule
\multirow[c]{3}{*}{SAPSO-rl} & CPU (\si{\joule}) & 3802712.12 & 124227.25 & 1116918.58 & 59046.54 & \textbf{63569.95} & 2657.91 & 1.73 & 0.08 & 429.78 & 21.01 & 4983632.16 & 157279.03 \\
 & DRAM (\si{\joule}) & 208849.89 & 18209.81 & 92110.50 & 1935.67 & 2239.98 & 217.23 & 0.10 & 0.01 & 26.75 & 1.41 & 303227.21 & 17051.99 \\
 & Time (\si{\second}) & 74549.27 & 2274.47 & 21455.75 & 452.93 & \textbf{1634.13} & 16.43 & 0.03 & 0.00 & 8.99 & 0.17 & 97648.18 & 2244.17 \\
 \midrule\midrule
 \multirow[c]{3}{*}{GA} & CPU (\si{\joule}) & 8506909.25 & 274586.56 & NA & NA & NA & NA & \textbf{1.40} & 0.12 & 1175.90 & 62.69 & 8508086.55 & 274631.35 \\
 & DRAM (\si{\joule}) & 471427.84 & 46411.70 & NA & NA & NA & NA & \textbf{0.07} & 0.02 & 164.04 & 19.88 & 471591.95 & 46430.64 \\
 & Time (\si{\second}) & 167419.86 & 4190.30 & NA & NA & NA & NA & 0.03 & 0.00 & 51.63 & 2.36 & 167471.53 & 4191.12 \\
\midrule
\multirow[c]{3}{*}{SAGA-ps} & CPU (\si{\joule}) & \textbf{54193.37} & 854.81 & \textbf{351.14} & 1.11 & 99870.36 & 5713.34 & 1.45 & 0.05 & 1136.85 & 38.00 & \textbf{155553.16} & 5293.69 \\
 & DRAM (\si{\joule}) & \textbf{2859.39} & 287.25 & \textbf{20.48} & 2.99 & 5560.22 & 1644.83 & 0.08 & 0.01 & 63.58 & 18.74 & \textbf{8503.75} & 1930.78 \\
 & Time (\si{\second}) & \textbf{1062.48} & 12.52 & \textbf{6.94} & 0.20 & 2154.72 & 119.31 & \textbf{0.03} & 0.00 & \textbf{22.11} & 0.84 & \textbf{3246.28} & 115.54 \\
\midrule
\multirow[c]{3}{*}{SAGA-pl} & CPU (\si{\joule}) & 2694260.09 & 103445.99 & 4260.66 & 1369.77 & 63609.61 & 2313.36 & 1.48 & 0.08 & 1116.22 & 43.36 & 2763248.06 & 105022.43 \\
 & DRAM (\si{\joule}) & 146687.30 & 15288.22 & 328.36 & 12.57 & \textbf{1994.59} & 347.95 & 0.08 & 0.02 & 69.21 & 4.64 & 149079.54 & 14995.80 \\
 & Time (\si{\second}) & 53037.90 & 1555.61 & 83.18 & 1.96 & 1650.55 & 14.96 & 0.03 & 0.00 & 30.94 & 0.86 & 54802.60 & 1561.95 \\
\midrule
\multirow[c]{3}{*}{SAGA-rs} & CPU (\si{\joule}) & 1126591.39 & 67990.56 & 351939.82 & 16880.59 & 89277.16 & 2456.07 & 1.47 & 0.07 & 1106.91 & 50.15 & 1568916.76 & 72643.68 \\
 & DRAM (\si{\joule}) & 64810.73 & 3528.77 & 24926.87 & 892.09 & 3175.20 & 186.30 & 0.09 & 0.01 & \textbf{40.23} & 1.85 & 92953.13 & 3806.01 \\
 & Time (\si{\second}) & 22112.33 & 1355.63 & 6607.90 & 115.63 & 2249.86 & 35.79 & 0.03 & 0.00 & 23.31 & 0.84 & 30993.43 & 1340.82 \\
\midrule
\multirow[c]{3}{*}{SAGA-rl} & CPU (\si{\joule}) & 3294572.73 & 122570.98 & 1114994.90 & 61473.61 & \textbf{63238.66} & 2600.85 & 1.45 & 0.10 & \textbf{1104.69} & 44.11 & 4473912.43 & 173238.29 \\
 & DRAM (\si{\joule}) & 181962.16 & 16418.83 & 91415.63 & 1902.56 & 2164.66 & 174.38 & 0.08 & 0.01 & 71.35 & 6.37 & 275613.88 & 16176.18 \\
 & Time (\si{\second}) & 64802.67 & 1892.68 & 21409.07 & 490.19 & \textbf{1627.46} & 14.28 & 0.03 & 0.00 & 30.72 & 0.94 & 87869.95 & 2136.86 \\
 \bottomrule
\end{tabular}
}
\vspace{-2em}
\end{table*}
\begin{table*}[tb]
\tabcolsep = 5.4pt
\caption{Paris instance: Average energy consumption, execution time, and their standard deviation for each optimization component}
\label{tab:measurement_summary_paris}
\vspace{-0.5em}
\centering
\scriptsize{
\begin{tabular}{llrrrrrrrrrrrr}
\toprule
 &  & \multicolumn{2}{c}{Evaluation} & \multicolumn{2}{c}{Training} & \multicolumn{2}{c}{Use} & \multicolumn{2}{c}{Initialization} & \multicolumn{2}{c}{Update} & \multicolumn{2}{c}{Total} \\
 \cmidrule(lr){3-4}\cmidrule(lr){5-6}\cmidrule(lr){7-8}\cmidrule(lr){9-10}\cmidrule(lr){11-12}\cmidrule(lr){13-14}
 & Measure & Avg. & Stdv. & Avg. & Stdv. & Avg. & Stdv. & Avg. & Stdv. & Avg. & Stdv. & Avg. & Stdv. \\
\midrule
\multirow[c]{3}{*}{PSO} & CPU (\si{\joule}) & 6687795.45 & 257303.00 & NA & NA & NA & NA & 1.75 & 0.11 & \textbf{429.63} & 10.77 & 6688226.83 & 257307.45 \\
 & DRAM (\si{\joule}) & 368305.83 & 30652.83 & NA & NA & NA & NA & 0.10 & 0.02 & 57.55 & 6.45 & 368363.48 & 30658.71 \\
 & Time (\si{\second}) & 131418.63 & 4081.36 & NA & NA & NA & NA & 0.03 & 0.00 & 14.18 & 0.44 & 131432.84 & 4081.36 \\
\midrule
\multirow[c]{3}{*}{SAPSO-ps} & CPU (\si{\joule}) & \textbf{33575.84} & 643.33 & \textbf{352.69} & 1.65 & 100383.96 & 5936.11 & 1.74 & 0.07 & 466.35 & 5.63 & \textbf{134780.58} & 5401.95 \\
 & DRAM (\si{\joule}) & \textbf{1923.80} & 167.79 & \textbf{21.49} & 3.52 & 5730.22 & 1726.27 & 0.10 & 0.01 & 26.26 & 7.46 & \textbf{7701.87} & 1831.27 \\
 & Time (\si{\second}) & \textbf{658.57} & 6.98 & \textbf{6.98} & 0.20 & 2166.21 & 116.03 & 0.03 & 0.00 & \textbf{7.07} & 0.22 & \textbf{2838.86} & 118.27 \\
\midrule
\multirow[c]{3}{*}{SAPSO-pl} & CPU (\si{\joule}) & 1898642.30 & 81128.39 & 4870.26 & 271.81 & 64278.98 & 2407.43 & \textbf{1.69} & 0.13 & 431.91 & 19.91 & 1968225.15 & 82884.68 \\
 & DRAM (\si{\joule}) & 107029.55 & 12962.31 & 361.47 & 11.64 & \textbf{2186.89} & 274.39 & \textbf{0.08} & 0.02 & 26.45 & 2.23 & 109604.44 & 12773.36 \\
 & Time (\si{\second}) & 37256.14 & 1259.79 & 85.41 & 1.79 & 1666.66 & 15.67 & 0.03 & 0.00 & 9.03 & 0.23 & 39017.27 & 1262.39 \\
\midrule
\multirow[c]{3}{*}{SAPSO-rs} & CPU (\si{\joule}) & 716905.21 & 27410.30 & 356143.64 & 17749.39 & 89670.58 & 2383.64 & 1.76 & 0.08 & 447.09 & 17.97 & 1163168.28 & 43544.92 \\
 & DRAM (\si{\joule}) & 42721.90 & 2349.59 & 26293.51 & 1051.80 & 3385.97 & 345.15 & 0.11 & 0.01 & \textbf{17.05} & 1.83 & 72418.52 & 3659.75 \\
 & Time (\si{\second}) & 14101.20 & 429.63 & 6703.78 & 101.97 & 2261.14 & 32.11 & \textbf{0.03} & 0.00 & 7.33 & 0.07 & 23073.47 & 477.89 \\
\midrule
\multirow[c]{3}{*}{SAPSO-rl} & CPU (\si{\joule}) & 2382360.10 & 100280.89 & 1139741.48 & 59331.73 & \textbf{63718.35} & 2545.81 & 1.75 & 0.06 & 433.74 & 19.87 & 3586255.42 & 145048.35 \\
 & DRAM (\si{\joule}) & 139516.61 & 14850.41 & 96083.57 & 2422.64 & 2247.18 & 199.57 & 0.10 & 0.01 & 27.60 & 2.96 & 237875.07 & 15182.02 \\
 & Time (\si{\second}) & 46700.05 & 1565.91 & 21828.70 & 443.11 & \textbf{1637.71} & 13.03 & 0.03 & 0.00 & 9.03 & 0.13 & 70175.53 & 1802.65 \\
 \midrule\midrule
\multirow[c]{3}{*}{GA} & CPU (\si{\joule}) & 5819731.42 & 112232.67 & NA & NA & NA & NA & 1.45 & 0.09 & 1208.34 & 45.07 & 5820941.20 & 112260.30 \\
 & DRAM (\si{\joule}) & 331642.18 & 37862.03 & NA & NA & NA & NA & \textbf{0.05} & 0.02 & 171.34 & 23.00 & 331813.57 & 37884.37 \\
 & Time (\si{\second}) & 114538.23 & 1295.78 & NA & NA & NA & NA & 0.03 & 0.00 & 53.12 & 1.79 & 114591.37 & 1295.41 \\
\midrule
\multirow[c]{3}{*}{SAGA-ps} & CPU (\si{\joule}) & \textbf{33550.48} & 591.00 & \textbf{352.16} & 5.96 & 99924.78 & 5740.21 & \textbf{1.46} & 0.13 & 1157.66 & 23.52 & \textbf{134986.54} & 5271.15 \\
 & DRAM (\si{\joule}) & \textbf{1798.12} & 250.77 & \textbf{20.33} & 3.40 & 5580.90 & 1849.81 & 0.07 & 0.03 & 65.10 & 20.68 & \textbf{7464.52} & 2079.58 \\
 & Time (\si{\second}) & \textbf{656.07} & 7.24 & \textbf{7.10} & 0.38 & 2155.71 & 119.53 & 0.03 & 0.00 & \textbf{22.52} & 0.72 & \textbf{2841.44} & 122.57 \\
\midrule
\multirow[c]{3}{*}{SAGA-pl} & CPU (\si{\joule}) & 1723804.65 & 37013.63 & 4429.94 & 1424.54 & 63631.21 & 2322.74 & 1.49 & 0.06 & 1123.49 & 37.36 & 1792990.79 & 39423.05 \\
 & DRAM (\si{\joule}) & 97991.70 & 11094.34 & 354.56 & 12.07 & \textbf{1979.66} & 348.25 & 0.09 & 0.01 & 71.54 & 5.34 & 100397.55 & 10810.42 \\
 & Time (\si{\second}) & 33907.04 & 434.68 & 85.00 & 1.69 & 1650.85 & 14.17 & 0.03 & 0.00 & 31.21 & 0.56 & 35674.13 & 437.08 \\
\midrule
\multirow[c]{3}{*}{SAGA-rs} & CPU (\si{\joule}) & 751777.22 & 34148.62 & 355343.12 & 15837.56 & 89134.49 & 2223.04 & 1.47 & 0.06 & \textbf{1110.62} & 29.33 & 1197366.92 & 50044.52 \\
 & DRAM (\si{\joule}) & 44479.93 & 1721.75 & 26174.96 & 847.14 & 3230.47 & 151.88 & 0.09 & 0.01 & \textbf{41.79} & 2.25 & 73927.24 & 2487.40 \\
 & Time (\si{\second}) & 14756.00 & 486.47 & 6689.10 & 103.29 & 2245.81 & 28.35 & \textbf{0.03} & 0.00 & 23.37 & 0.67 & 23714.31 & 582.48 \\
\midrule
\multirow[c]{3}{*}{SAGA-rl} & CPU (\si{\joule}) & 2138857.86 & 48332.43 & 1134596.65 & 59725.84 & \textbf{63259.98} & 2719.56 & 1.50 & 0.10 & 1134.36 & 52.70 & 3337850.36 & 104227.24 \\
 & DRAM (\si{\joule}) & 123823.49 & 10262.28 & 94878.83 & 1501.61 & 2162.69 & 151.01 & 0.08 & 0.01 & 75.22 & 5.90 & 220940.31 & 9892.74 \\
 & Time (\si{\second}) & 42055.02 & 561.84 & 21774.67 & 478.33 & \textbf{1627.88} & 16.28 & 0.03 & 0.00 & 31.62 & 0.79 & 65489.21 & 832.17 \\
\bottomrule
\end{tabular}
}
\vspace{-2em}
\end{table*}

This experiment aims to compare the computational performance of SAPSO and SAGA in terms of energy consumption, execution time, and memory usage. The following subsections first examine energy consumption and execution time, followed by an analysis of memory usage, and finally provide a summary.

\subsubsection{Energy Consumption and Time}

Tables~\ref{tab:measurement_summary_malaga}--\ref{tab:measurement_summary_paris} present the average energy consumption and execution time. The values are broken down by algorithm component, and overall totals are displayed in the ``Total'' column. For each metric, the minimum value within each algorithm family is highlighted in bold. Note that the baseline PSO and GA, which do not utilize a surrogate model, have no NN training or usage; these entries are marked as ``NA.''

Regarding energy consumption of the PSO families, the baseline PSO spends most of its energy on solution evaluations via the simulator, while initialization and update operations consume only a negligible amount. SAPSO families follow a similar pattern but introduce additional costs associated with surrogate training and usage. Among them, SAPSO-ps and SAPSO-pl---both of which use pre-training---consume less energy for training than SAPSO-rs and SAPSO-rl, which perform retraining during optimization. SAPSO-ps achieves the lowest training cost due to the use of a smaller dataset.

In terms of surrogate usage, SAPSO-ps and SAPSO-rs consume more energy than SAPSO-pl and SAPSO-rl. This is because the former begins using the surrogate earlier in the search process, resulting in more frequent usage. Notably, SAPSO-rs, despite using a small dataset, benefits from retraining, which improves prediction efficiency and reduces energy per usage, as discussed in Section~\ref{sec:NN_use}. Overall, all SAPSO families significantly reduce total energy consumption compared to the baseline PSO, demonstrating the effectiveness of surrogate models in lowering the computational cost of solution evaluations.

Execution time shows a similar trend. The baseline PSO takes the longest time due to repeated expensive evaluations. All SAPSO families complete faster, with SAPSO-ps---a pre-trained variant---being the fastest. Retraining-based algorithms, particularly SAPSO-rl, require more time due to repeated surrogate model updates. The time spent on surrogate usage remains relatively small across all configurations. SAPSO-ps and SAPSO-rs show slightly longer surrogate usage time due to more frequent invocations, but as confirmed in Section~\ref{sec:NN_use}, the time per evaluation remains stable regardless of dataset size.

Similar to PSO, the baseline GA spends most of its energy and time on solution evaluations using the actual simulator, while initialization and update operations incur minimal costs. All SAGA families reduce total energy consumption compared to the baseline GA, indicating that surrogate integration is effective across both algorithmic families.

Regarding surrogate training, SAGA-ps and SAGA-pl consume less energy than SAGA-rs and SAGA-rl. This is because the pre-training algorithms perform training only once, whereas the retraining-based algorithms incur repeated overhead. In surrogate usage, SAGA-ps and SAGA-rs consume more energy than their large-dataset counterparts (SAGA-pl and SAGA-rl), as they begin using the surrogate earlier in the search. Notably, SAGA-rs shows reduced energy consumption per use compared to SAGA-ps, suggesting improved prediction efficiency due to retraining, as also observed in SAPSO-rs.

Execution time follows the same general pattern. All SAGA families complete faster than the baseline GA. Among them, SAGA-ps and SAGA-pl are the fastest, while SAGA-rs and SAGA-rl require more time due to repeated retraining. The time spent on surrogate usage remains small in all configurations. SAGA-ps and SAGA-rs exhibit slightly longer usage time due to more frequent calls, but as with the PSO families, the time per evaluation remains consistent across different dataset sizes.

\subsubsection{Memory Usage}
\begin{table}[!tb]
\centering
\caption{Average memory usage (\si{\mega\byte}) and their standard deviation}
\vspace{-0.5em}
\label{tb:measurement_summary_memory}
\scriptsize{
\begin{tabular}{lrrrrrr}
\toprule
 & \multicolumn{2}{c}{\Malaga} & \multicolumn{2}{c}{Stockholm} & \multicolumn{2}{c}{Paris} \\
 & Avg. & Stdv. & Avg. & Stdv. & Avg. & Stdv. \\
\midrule
PSO & \textbf{1.59} & 0.00 & \textbf{2.55} & 0.01 & \textbf{2.60} & 0.00 \\
SAPSO-ps & 21.22 & 0.00 & 22.19 & 0.00 & 22.23 & 0.02 \\
SAPSO-pl & 66.70 & 1.75 & 124.73 & 1.71 & 127.63 & 1.52 \\
SAPSO-rs & 590.86 & 1.53 & 612.03 & 1.80 & 612.99 & 1.75 \\
SAPSO-rl & 497.73 & 2.15 & 568.89 & 2.61 & 572.37 & 2.35 \\
\midrule
GA & \textbf{1.00} & 0.00 & \textbf{1.41} & 0.00 & \textbf{1.43} & 0.01 \\
SAGA-ps & 20.57 & 0.00 & 20.99 & 0.00 & 21.01 & 0.00 \\
SAGA-pl & 53.35 & 1.88 & 99.37 & 1.82 & 100.70 & 1.88 \\
SAGA-rs & 585.51 & 1.50 & 602.46 & 1.82 & 603.17 & 1.79 \\
SAGA-rl & 481.82 & 0.88 & 539.46 & 1.71 & 542.76 & 1.18 \\
\bottomrule
\end{tabular}
}
\vspace{-2em}
\end{table}
Table~\ref{tb:measurement_summary_memory} reports the average memory usage and standard deviation. The minimum value within each algorithm family is highlighted in bold. Note that since tracemalloc does not capture memory usage of subprocesses, the reported memory usage reflects only the memory consumed by search algorithms, excluding that of the SUMO simulator.

Among these, the baseline PSO consistently exhibits the lowest memory consumption. The standard deviations are nearly zero, indicating highly stable memory allocation throughout execution. In contrast, all SAPSO families require more memory due to the overhead introduced by surrogate integration. Even SAPSO-ps, which performs one-time training with a small dataset and uses the surrogate exclusively thereafter, consumes more memory than the baseline. This is primarily due to the memory needed to store the surrogate model and its associated data structures, as noted in Section~\ref{sec:NN_training}.

SAPSO-rs and SAPSO-rl show the highest memory usage. SAPSO-rs, in particular, uses the most memory among all configurations, followed by SAPSO-rl. These retraining-based approaches increase memory demand because they must retain solution archives and repeatedly construct and manage NN models throughout the search.

A similar pattern is observed for GA families: the baseline GA has the lowest memory usage, and all SAGA families require more. SAGA-ps shows the lowest usage among the SAGA families, followed by SAGA-pl. Meanwhile, SAGA-rs and SAGA-rl consume substantially more memory due to retraining, mirroring the behavior seen in SAPSO.

Across all instances, memory usage increases from \Malaga to Stockholm to Paris. This trend holds for both PSO and GA families, and is particularly pronounced in surrogate-assisted configurations. This suggests that problem complexity or dimensionality directly affects memory requirements.

Comparing PSO and GA families more broadly, GA consistently requires less memory under all configurations. This difference arises from the intrinsic characteristics of the algorithms: PSO must manage both position and velocity vectors for each particle, whereas GA only handles discrete solution representations, leading to lower memory usage overall.

\subsubsection{Summary}
The results demonstrate that surrogate-assisted algorithms can significantly reduce energy consumption and execution time, although at the cost of substantially increased memory usage. The detailed results are summarized as follows:
\begin{itemize}
    \setlength\parskip{-1em}%
    \setlength\itemsep{0em}%
\item \textbf{Pre-training on small datasets} (e.g., SAPSO-ps, SAGA-ps):
Achieve the largest reductions in energy and time, exceeding 95\%. However, memory usage increases by approximately 9--13 times for SAPSO-ps and 14--20 times for SAGA-ps compared to the baselines.
    \setlength\parskip{0em}%
\item \textbf{Pre-training on large datasets} (e.g., SAPSO-pl, SAGA-pl):
Reduce energy and time by approximately 70\%, while memory usage increases by around 42--49 times for SAPSO-pl and 43--70 times for SAGA-pl.
\item \textbf{Retraining on small datasets} (e.g., SAPSO-rs, SAGA-rs):
Achieve up to 85\% reductions in energy and time. In contrast, memory usage increases significantly, by approximately 236--371 times for SAPSO-rs and 422--586 times for SAGA-rs.
\item \textbf{Retraining on large datasets} (e.g., SAPSO-rl, SAGA-rl):
Provide the smallest improvements in energy and time, with reductions of around 30–50\%, accompanied by memory usage increases of approximately 220--312 times for SAPSO-rl and 379--382 times for SAGA-rl.
\end{itemize}
\vspace{-1em}

Overall, these results highlight a clear trade-off: while SAEAs effectively reduce energy consumption and execution time, especially with small pre-training datasets, they incur considerable memory overhead, which must be carefully considered in practical applications.

The next subsection focuses on analyzing the optimization accuracy of each surrogate-assisted algorithm.

\subsection{Experiment 2: Optimization Accuracy of the Algorithm}
\begin{figure}[!tb]
    \centering
    \begin{minipage}[t]{0.49\columnwidth}
    \includegraphics[width=\textwidth]{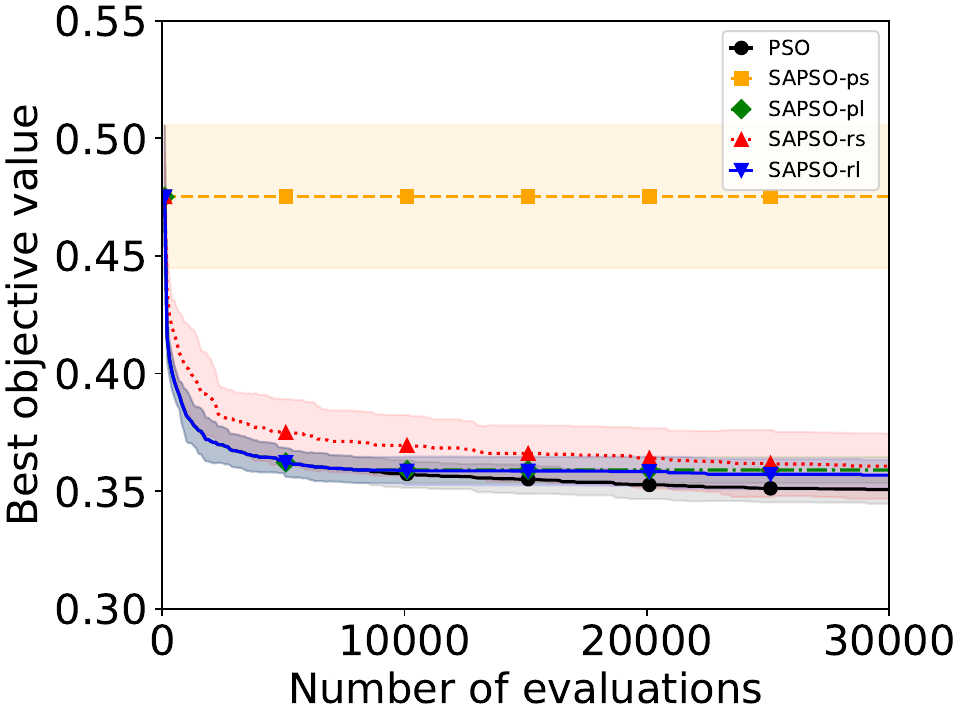}
    \subcaption{PSO families}
    \end{minipage}
    \hfill
    \centering
    \begin{minipage}[t]{0.49\columnwidth}
    \includegraphics[width=\textwidth]{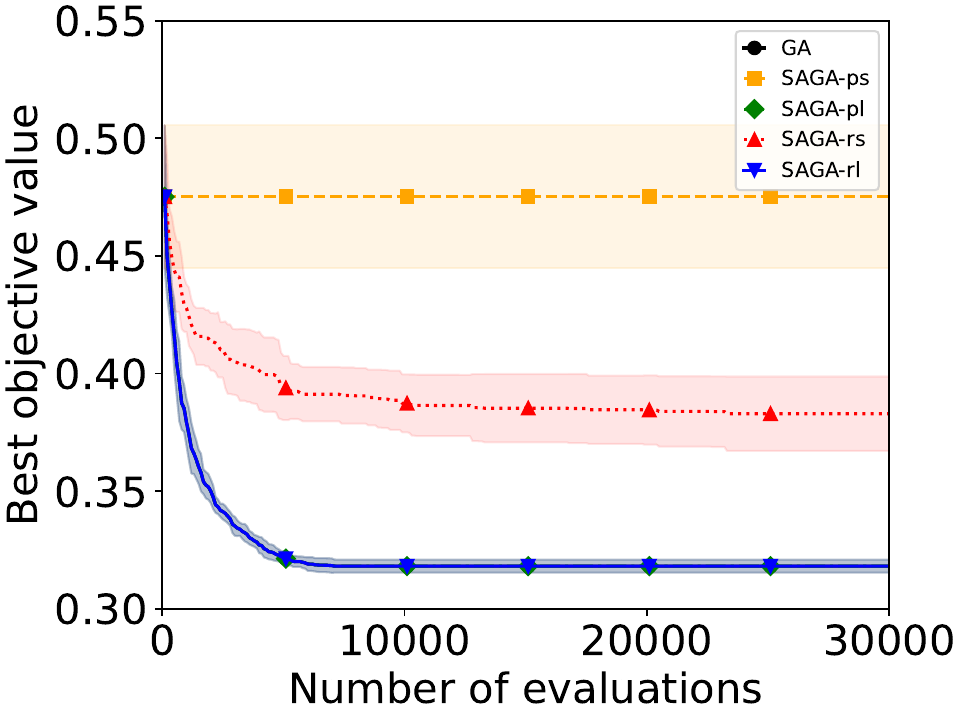}
    \subcaption{GA families}
    \end{minipage}
\vspace{-0.5em}
    \caption{Average objective values with standard deviation for \Malaga}
    \label{fig:evaluation_best_fit}
    \vspace{-1.5em}
\end{figure}
Fig.~\ref{fig:evaluation_best_fit} illustrates the transition of the average best objective values obtained during the optimization process for the \Malaga instance. The horizontal axis shows the number of actual evaluations, while the vertical axis indicates the average best objective values across all runs. Although surrogate-assisted algorithms internally use predicted objective values to guide the search, only the results from actual evaluations are reported here to ensure a fair comparison of optimization performance. Due to space constraints, we only provide the results of \Malaga, but similar trends were observed in the other instances.

A notable observation is that SAPSO-ps and SAGA-ps exhibit little to no improvement in best objective values beyond the initial population. A similar pattern is observed in SAPSO-pl: although some progress is made early in the optimization, the search stagnates soon after the surrogate model is introduced at 8200 evaluations. These trends suggest that the pre-training families, once the surrogate model is in use, do not lead to further improvements in solution quality---highlighting the limitations of static surrogates trained on fixed datasets.

In contrast, the retraining-based SAPSO families show a more consistent improvement. SAPSO-rs begins using the surrogate after 100 initial evaluations and continues to discover better solutions throughout the search. Likewise, SAPSO-rl activates the surrogate after 8200 evaluations and still outperforms SAPSO-pl, which does not include retraining.

The behavior of the retraining-based SAGA families is different. SAGA-rs achieves slight improvements after switching to surrogate-based optimization at 100 evaluations; however, the progress is slower than in SAPSO-rs, and the overall gains are limited. This suggests that the GA search may lack the diversity needed to fully leverage surrogate retraining. In the case of SAGA-rl, the search converges before the surrogate is activated, and no further improvements are observed thereafter.

\begin{figure}[!tb]
    \centering
    \includegraphics[width=0.3\textwidth]{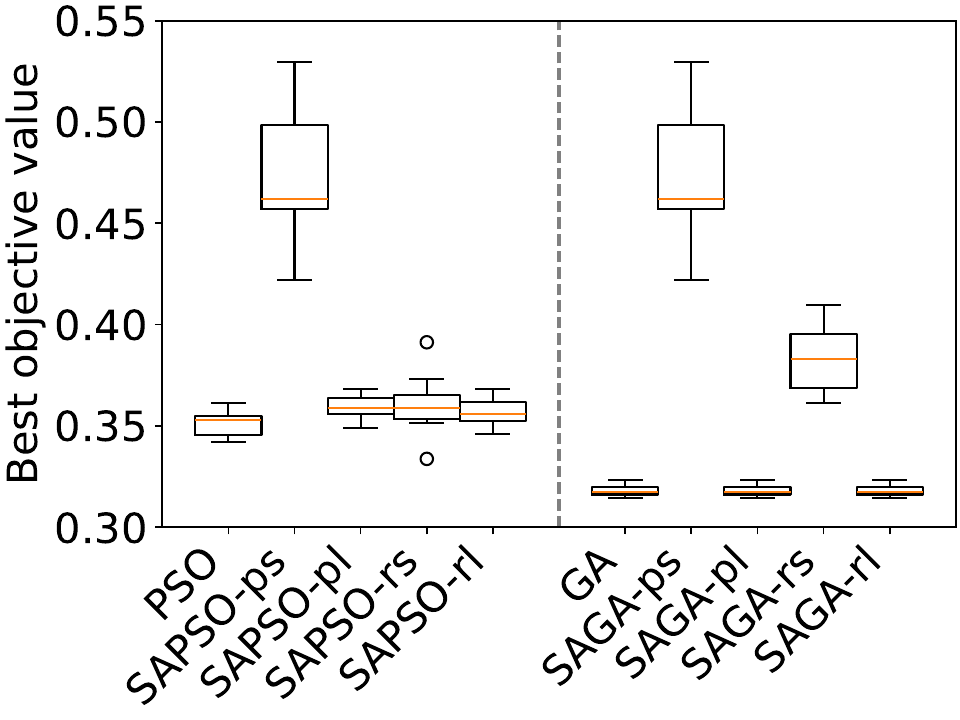}
\vspace{-0.5em}
    \caption{Best objective values for \Malaga (smaller is better)}
    \label{fig:final_fit}
    \vspace{-1.5em}
\end{figure}
Fig.~\ref{fig:final_fit} presents boxplots of the best objective values achieved by each algorithm for \Malaga. The horizontal axis shows the algorithm, while the vertical axis represents the best objective values obtained by each algorithm. The left five plots show the PSO families, while the right ones show the GA families.

Among the PSO families, SAPSO-ps shows the worst performance. SAPSO-pl and SAPSO-rl trained on large datasets yield similar objective values, indicating that retraining offers little advantage when the surrogate is already well-trained. In contrast, a clear gap is observed between SAPSO-ps and SAPSO-rs: with a small initial dataset, retraining significantly improves performance. Notably, SAPSO-rs achieves comparable results to SAPSO-pl, SAPSO-rl, and the baseline PSO.

A similar trend is observed in GA families. SAGA-ps performs poorly, while SAGA-pl and SAGA-rl reach performance levels similar to the baseline GA. However, SAGA-rs, despite retraining, performs worse than the other GA-based algorithms. This suggests that retraining from small datasets is less effective—or even detrimental—in GA.

Overall, GA families outperform PSO families across all conditions. This aligns with findings from~\citep{GARCIANIETO2012274}, where GA achieved better results than PSO due to its more aggressive convergence, enabled by faster diversity reduction.

However, in surrogate-assisted settings, this same behavior may limit the benefits of retraining. Our results indicate that GA’s rapid loss of diversity reduces opportunities to exploit updated surrogate predictions. In contrast, PSO’s ability to maintain higher diversity allows it to continue improving after surrogate activation, suggesting that moderate diversity is crucial for making retraining effective.

\begin{figure}[!tb]
    \centering
    \begin{minipage}[t]{0.49\columnwidth}
    \includegraphics[width=\textwidth]{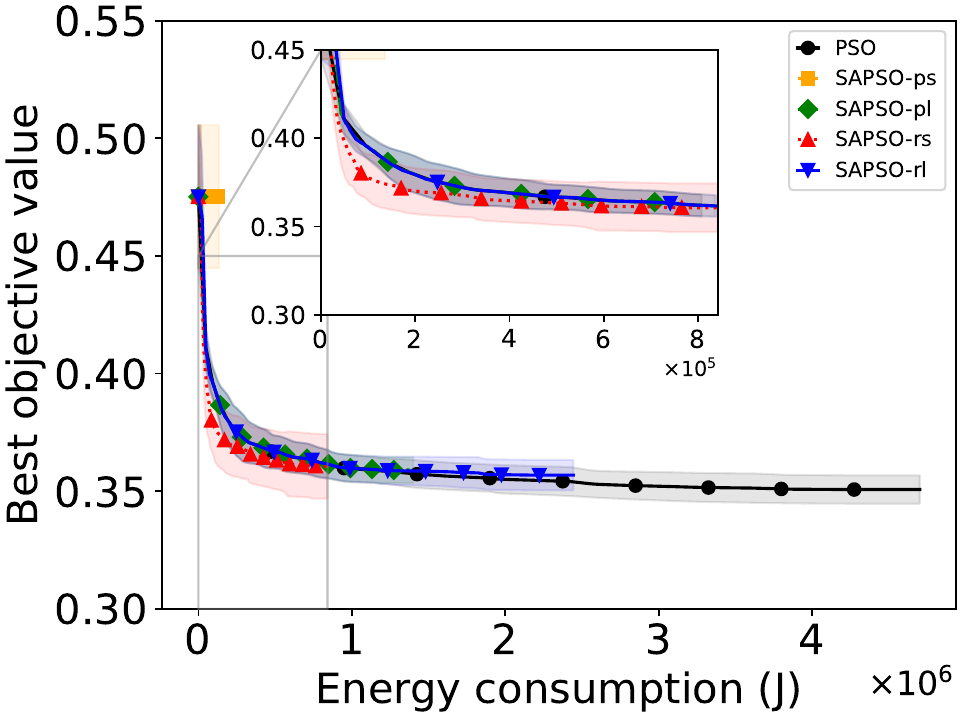}
    \subcaption{PSO families}
    \end{minipage}
    \hfill
    \begin{minipage}[t]{0.49\columnwidth}
    \includegraphics[width=\textwidth]{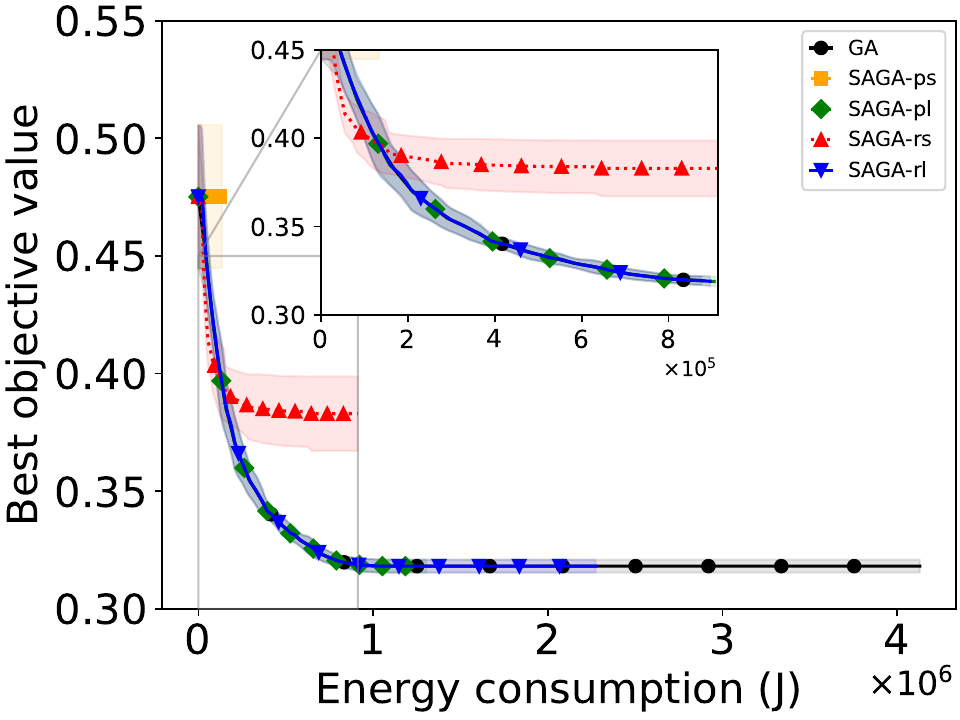}
    \subcaption{GA families}
    \end{minipage}
    \vspace{-0.5em}
    \caption{Best objective values over energy (\si{\joule}) on the \Malaga instance}
    \label{fig:energy_best_fit}
    \vspace{-1.5em}
\end{figure}
Fig.~\ref{fig:energy_best_fit} shows the evolution of best objective values via cumulative energy consumption for PSO and GA families.

Among the PSO families, SAPSO-ps exhibits early stagnation with no notable improvement after the initial phase. SAPSO-pl follows a trajectory similar to the baseline PSO in the early stages and achieves comparable final results while consuming less energy: approximately $1.5\times 10^{6}$~\si{\joule}. SAPSO-rl also mirrors PSO in the early phase, but shows slightly inferior final performance despite higher energy use: around $2.5\times 10^{6}$~\si{\joule}.

In contrast, SAPSO-rs achieves steady improvements throughout the search while using significantly less energy. When compared with PSO at equivalent energy levels (see zoomed-in views), SAPSO-rs consistently yields better average objective values across all instances, demonstrating highly efficient optimization under constrained energy budgets.

Turning to the GA families, SAGA-ps shows similar stagnation to SAPSO-ps. The other configurations all exhibit nearly identical trajectories, reaching their final objective values early in the search around $1.0\times 10^{6}$~\si{\joule}. This indicates that their searches converge before algorithmic differences take effect.

SAGA-rs follows a distinct pattern: It achieves a sharp drop in objective value at a very early stage, outperforming all other algorithms under tight energy constraints. Specifically, in the range of approximately $1.0\times 10^{5}$~\si{\joule}, SAGA-rs achieves the best results. However, its performance quickly plateaus, and its final objective values remain worse than those of other algorithms.

Comparing PSO and GA families, SAPSO-rs outperforms all GA configurations under tight energy budgets. This contrasts with the trend in Fig.~\ref{fig:final_fit}, where GA families outperform PSO ones in final performance. The reversal suggests that while GA families achieve better long-term solutions, SAPSO-rs is more energy-efficient in the early stages, making it preferable under strict energy constraints.

Based on these observations, the performance characteristics of SAEAs can be summarized as follows:

\begin{itemize}
    \setlength\parskip{-1em}%
    \setlength\itemsep{0em}%
\item \textbf{Pre-training on small datasets} (e.g., SAPSO-ps, SAGA-ps):
Exhibit little to no improvement beyond the initial population, indicating that static surrogates trained on limited data are insufficient for continued search progress.
    \setlength\parskip{0em}%
\item \textbf{Pre-training on large datasets} (e.g., SAPSO-pl, SAGA-pl):
Follow similar trajectories to their baseline counterparts and reach final objective values early in the search. Energy efficiency is improved compared to the baseline, especially in SAPSO-pl.
\item \textbf{Retraining on small datasets} (e.g., SAPSO-rs, SAGA-rs):
Show rapid objective value improvements in the early stages under tight energy budgets. SAPSO-rs achieves high efficiency. However, SAGA-rs tends to stagnate quickly, likely due to reduced population diversity.
\item \textbf{Retraining on large datasets} (e.g., SAPSO-rl, SAGA-rl):
Demonstrate only marginal improvements over their pre-trained counterparts. The benefit of retraining is limited when the surrogate is already well-trained.
\end{itemize}
\vspace{-1em}

Overall, PSO families tend to benefit more from surrogate retraining, possibly due to higher population diversity, which helps sustain exploration. In contrast, GA families show less improvement after surrogate activation, especially when diversity declines too early in the search.

Finally, the following subsection analyzes the prediction accuracy of the surrogate models used during the optimization.

\subsection{Experiment 3: Prediction Accuracy of the Surrogate}
\begin{figure}[!tb]
\centering
\begin{minipage}[t]{0.49\columnwidth}
   \includegraphics[width=\linewidth]{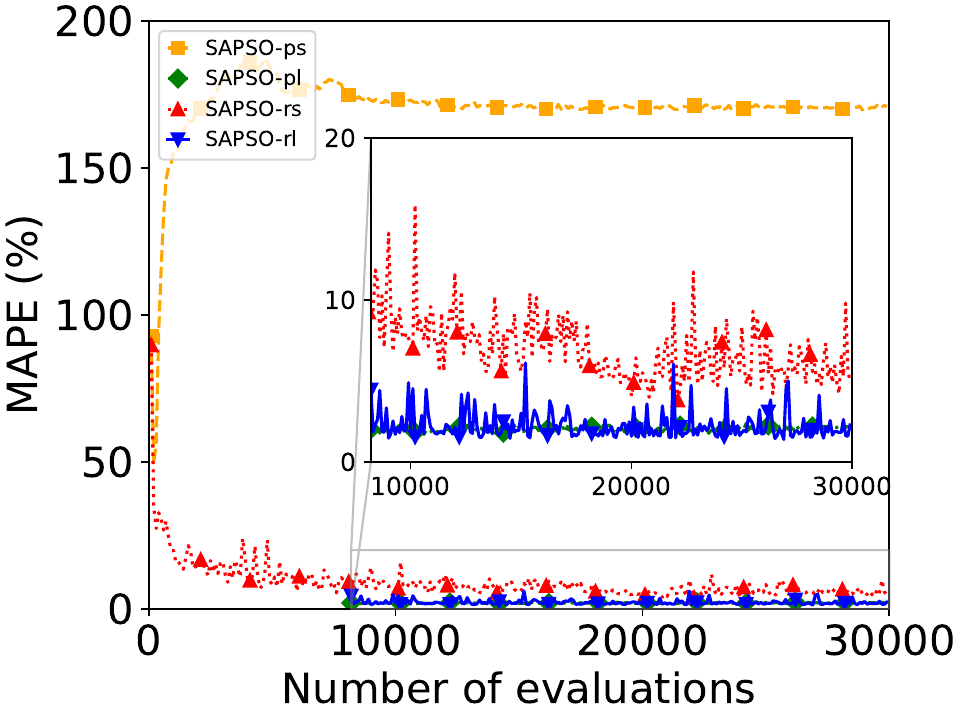}
   \subcaption{SAPSO families}\label{fig:MAPE_sapso}
\end{minipage}
\hfill
\begin{minipage}[t]{0.49\columnwidth}
   \includegraphics[width=\linewidth]{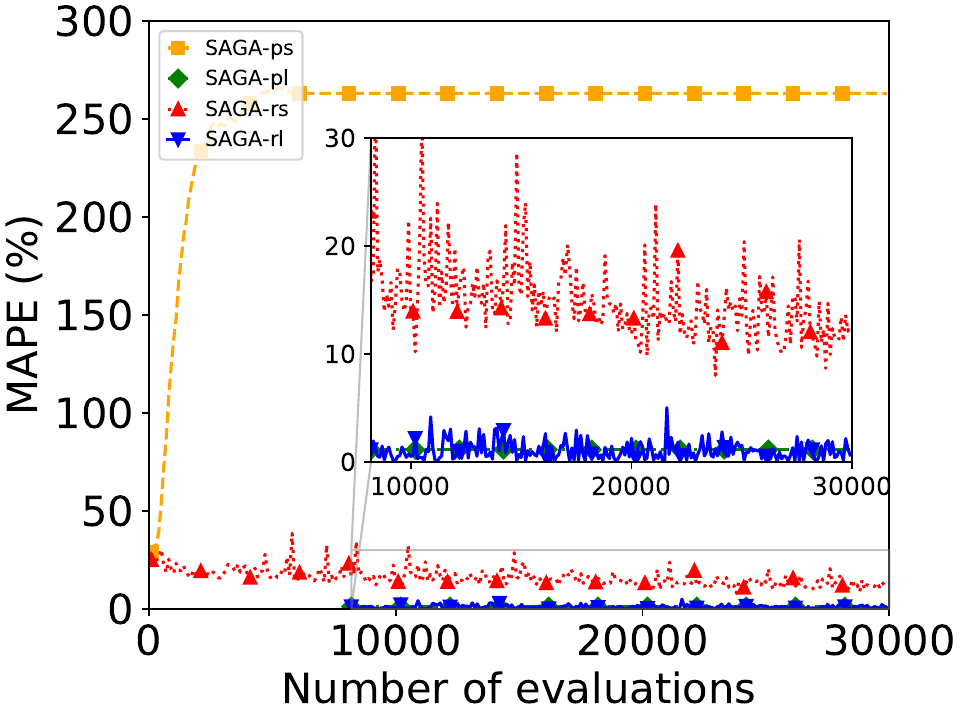}
   \subcaption{SAGA families}\label{fig:MAPE_saga}
\end{minipage}
\vspace{-0.5em}
\caption{MAPE via optimization for the \Malaga instance (lower is better)}
\label{fig:MAPE}
\vspace{-1.5em}
\end{figure}
To assess the prediction accuracy of surrogate models during the algorithm runs, Fig.~\ref{fig:MAPE} illustrates the transition of MAPE for SAPSO and SAGA families, respectively. During optimization, predicted objective values are used, but the accuracy of these predictions is independently assessed using actual objective values evaluated after the optimization. The horizontal axis represents the number of evaluations, and the vertical axis shows the corresponding MAPE. For each algorithm, the trial yielding the median final objective was selected for analysis.

Among the pre-training-based algorithms, SAPSO-ps and SAGA-ps exhibit a rapid increase in MAPE during the early stages of the search, which remains high throughout the process. This suggests that surrogate models trained on small datasets tend to generalize poorly and are unable to provide accurate predictions in previously unexplored regions.
By contrast, SAPSO-pl and SAGA-pl, which use large pre-trained datasets, maintain consistently low MAPE after surrogate usage begins at 8200 evaluations.

SAPSO-rs, which combines small initial datasets with retraining, demonstrates relatively stable and low MAPE. Although it performs slightly worse than large-dataset configurations, its accuracy improves steadily over time and approaches that of SAPSO-pl and SAPSO-rl. In contrast, SAGA-rs also maintains low MAPE throughout but fails to exhibit any improvement. This suggests premature convergence of the GA search, resulting in insufficient generation of new solutions and limited surrogate updates. Consequently, the surrogate model remains under-trained, and the prediction quality stagnates.

SAPSO-rl maintains a low MAPE. This indicates that frequent updates with a sufficiently rich dataset enable the surrogate to adapt to evolving search regions effectively. In contrast, SAGA-rl also achieves MAPE levels around 1\%, but further analysis reveals that it continues to evaluate nearly identical solutions with similar objective values. This suggests that the GA search had already converged when surrogate usage began, resulting in limited exploration and redundancy in predictions. The low MAPE in this case reflects accurate estimation of repeated solutions rather than effective guidance in the search.

These results underscore the importance of aligning surrogate modeling strategies with both the dynamics of the search process and the characteristics of the target problem. Pre-trained models based on small datasets tend to suffer from generalization failure and early stagnation. Those trained on larger datasets offer more robust initial accuracy but may degrade when used statically in evolving search regions. Retraining improves adaptability---particularly in PSO-based methods---but training is only effective when the optimization algorithm continues to generate diverse solutions; otherwise, as seen in some GA families, even adaptive surrogates provide limited benefit. Moreover, problem complexity strongly influences surrogate accuracy: as the search space becomes more challenging, model flexibility and continual adaptation become increasingly essential for maintaining reliable guidance.

\section{Conclusion}\label{sec:conclusions}

This paper investigates the use of deep NN surrogates as an efficient alternative to costly fitness function evaluations in optimization processes. To achieve this, we conducted extensive experiments comparing different surrogate training strategies—namely, state-of-the-art pre-training versus continuous (iterative) retraining—across various training dataset sizes. The experiments were designed to quantify the impact of surrogate models on key computational metrics, including energy consumption, execution time, and memory usage, providing a holistic evaluation of their performance in surrogate-assisted metaheuristics. Let us start the section by quickly summarising our main conclusions: 
\begin{itemize}
    \setlength\parskip{-1em}%
    \setlength\itemsep{0em}%
\item We provide a wider characterization of surrogate-assisted algorithms in terms of energy, time, and memory.
    \setlength\parskip{0em}%
\item In solution evaluations, energy consumption and execution time are highly correlated, following a log-normal distribution, while memory usage shows minimal variation and no apparent relation to instance complexity.
\item Surrogate-assisted algorithms significantly reduce evaluation time, but the trade-off in energy and memory usage depends on the training strategy.
\item SAPSO-r (retrained surrogate) balances accuracy and energy efficiency, while SAPSO-p (pre-trained surrogate) is faster but may require retraining for accurate predictions.
\item The best algorithm varies by problem complexity: PSO without a surrogate is highly inefficient in large-scale problems due to excessive evaluations, while SAPSO-r performs best when energy constraints are critical.
\item All the research questions are answered, confirming that surrogates consume non-negligible energy, efficiency must be redefined beyond runtime, and adaptive retraining of surrogates is often preferable.
\item An unexpected result is that larger training datasets lead to lower energy during inference as sparsity in NN increases.
\end{itemize}
\vspace{-1em}

We now summarise our conclusions in a topic-wise manner. Let us start with the characterization of both algorithms and surrogates in terms of time, memory, and energy. The study has thoroughly characterized different optimization algorithms, with and without surrogates, in terms of execution time, memory usage, and energy consumption. The results confirm that execution time and energy consumption are closely correlated, following a log-normal distribution across all problem instances. While surrogates drastically reduce execution time by replacing expensive fitness function evaluations, they introduce a trade-off in terms of energy due to model training and inference. Memory consumption, in contrast, remains relatively constant regardless of the optimization strategy, primarily dictated by problem dimensionality rather than algorithmic choice. These findings suggest that the selection of an optimization approach should not be based solely on runtime but rather on a holistic assessment of all three computational metrics.

As to the best algorithms, the experimental results indicate that no single algorithm dominates across all metrics, but instead, clear patterns emerge depending on the optimization trade-offs. Traditional PSO without a surrogate is highly energy-intensive, as it repeatedly evaluates the expensive real function, leading to exponential energy growth with problem size. In contrast, SAPSO-p (pre-trained surrogate) significantly reduces execution time but at the cost of an initial training overhead. Finally, SAPSO-r (retrained surrogate), while requiring additional energy for periodic retraining, provides a balanced approach, optimizing time, memory, and energy across different problem scales. This suggests that, in computationally constrained environments, an adaptive retraining strategy (SAPSO-r) is preferable, whereas for scenarios with fixed problem structures, pre-trained surrogates (SAPSO-p) may suffice.

As to the research questions answered in this article, we think that this study successfully answers all of them. The results clearly show that surrogate models consume non-trivial energy, sometimes offsetting the savings from reduced function evaluations. This demonstrates that energy efficiency must be redefined beyond simple runtime reductions, incorporating energy profiles into surrogate-assisted algorithm evaluation. The findings also confirm that incremental retraining (SAPSO-r) often outperforms a fixed pre-trained model (SAPSO-p), striking a better balance between accuracy, execution time, and energy consumption. Finally, we faced the question of whether a perfectly trained surrogate is necessary, concluding that a well-calibrated but not over-trained NN can be highly effective, as excessive training does not always yield significant gains in optimization performance.

An unexpected but highly relevant result is that larger training datasets for the surrogate lead to lower energy consumption during inference. This is due to the observed increase in sparsity within the trained NNs, where higher training volumes cause a greater proportion of neurons to become inactive (ReLU neurons outputting zeros). This sparsity effect leads to reduced computational complexity and, consequently, lower energy consumption during fitness evaluations. Additionally, the results confirm that problem complexity (number of intersections, simulation time, and number of vehicles) has a direct impact on the relative efficiency of surrogate-based optimization, suggesting that problem-aware surrogate selection strategies could further improve energy-aware optimization.

As a final conclusion, this study presents a novel energy-aware perspective on surrogate-assisted metaheuristics, emphasizing that while surrogates improve efficiency, their computational cost must be carefully assessed. By evaluating time, memory, and energy holistically, the study provides clear guidelines for selecting surrogate models based on problem characteristics. The findings contribute to the growing field of green computing and energy-efficient AI, offering practical insights for researchers and practitioners optimizing real-world problems where computational cost is a key concern. Future work could extend this methodology to other metaheuristics and explore alternative surrogate architectures to further optimize the trade-off between accuracy, efficiency, and sustainability.

Our findings underscore significant benefits of adopting surrogate models. (1) In terms of energy, the use of a well-trained pre-trained surrogate resulted in reductions of up to 98\% compared to the conventional fitness function evaluations. (2) Execution time improvements were equally compelling, with reductions of approximately 98\%, thereby accelerating the overall optimization process. (3) Memory usage also saw notable decreases, reaching reductions of around 99\%, which underscores the efficiency gains in resource allocation. Collectively, these results demonstrate that while both training strategies enhance performance, pre-training offers a more advantageous balance between efficiency and accuracy, making it a promising approach for large-scale surrogate-assisted optimization.

Overall, our study states the need for an energy-aware approach to evaluate surrogate-assisted metaheuristics, offering practical recommendations for solving real-world problems.

Despite the comprehensive scope of this study, several limitations remain. First, we focused solely on fully connected NNs; future work should explore other architectures (e.g., convolution, recurrent, transformers) with potentially different efficiency–accuracy trade-offs. Second, experiments were limited to traffic light scheduling. While practically relevant, broader validation is needed to assess generalizability to other problems. Third, energy measurements were performed in a controlled hardware setting; therefore, results may vary across different platforms. Future directions include applying this framework to other metaheuristics, testing diverse machine learning models, and investigating online adaptation and energy-aware scheduling in dynamic environments.

\section*{CRediT authorship contribution statement}
\textbf{Tomohiro Harada: } Writing -- original draft, Conceptualization, Investigation, Methodology, Formal analysis, Software, Visualization, Funding Acquisition. \textbf{Enrique Alba: } Writing -- review \& editing, Conceptualization, Investigation, Methodology, Supervision, Funding Acquisition. \textbf{Gabriel Luque} Writing -- review \& editing, Methodology, Investigation, Software, Funding Acquisition.
\section*{Declaration of Generative AI and AI-assisted technologies in the writing process}
During the preparation of this work, the authors used Word Processing tools, Spreadsheets, and AI assistants in order to improve the readability of the text and figures. After that, the authors reviewed and edited the content as needed and then took full responsibility for the published article.
\section*{Declaration of competing interest}
The authors declare that they have no known competing interests or personal relationships that could have appeared to influence the work reported in this paper. This work does not represent the interest or opinion of any affiliated institution.

\section*{Acknowledgments}
This work was supported by the Japan Society for the Promotion of Science Grant-in-Aid for Young Scientists (Grant No. 21K17826) and the Okawa Foundation for Information and Telecommunications Research Grant 2024 (Grant No. 24-05). 

\section*{Disclaimer}
This paper does not represent the opinion of any of our affiliated organizations.

\bibliographystyle{elsarticle-num} 
\bibliography{references}






\end{document}